\documentclass[twocolumn,a4paper]{article}
\usepackage[utf8]{inputenc}
\DeclareUnicodeCharacter{2212}{-}
\pdfoutput=1

\usepackage[english]{babel}
\usepackage{lipsum}

\usepackage[T1]{fontenc}
\usepackage{newtxtext}

\usepackage{graphicx}
\graphicspath{{figures/}}

\usepackage{amsmath,amssymb,amsfonts}

\usepackage[numbers]{natbib}
\usepackage{authblk}
\usepackage{algorithmic}
\usepackage[hyphens]{url}
\usepackage{hyperref}
\usepackage{enumitem}
\usepackage{pbox}
\usepackage{titlesec}
\usepackage{indentfirst}
\usepackage{textcomp}
\usepackage{colortbl}
\usepackage{multirow}
\usepackage{pifont}

\newcommand{\N}{\ding{55}}
\newcommand{\Y}{\ding{51}}

\definecolor{TableGray}{gray}{0.925}

\setlist[itemize]{noitemsep}
\titlelabel{\thetitle.\hspace{0.5em}}

\newcommand{\insetfont}{\sffamily \footnotesize}

\begin{document}

\title{OG-SGG: Ontology-Guided Scene Graph Generation.
A Case Study in Transfer Learning for Telepresence Robotics}
\author[1]{Fernando Amodeo \thanks{famozur@upo.es}}
\author[2]{Fernando Caballero \thanks{fcaballero@us.es}}
\author[3]{Natalia Díaz-Rodríguez \thanks{nataliadiaz@ugr.es}}
\author[1]{Luis Merino \thanks{lmercab@upo.es}}
\affil[1]{Service Robotics Laboratory, Universidad Pablo de Olavide, Seville, Spain.}
\affil[2]{Service Robotics Laboratory, Universidad de Sevilla, Spain.}
\affil[3]{Computer Science and Artificial Intelligence Dept., Andalusian Research Institute in Data Science and Computational Intelligence (DaSCI), University of Granada, Spain}
\date{}
\maketitle

{\noindent\textbf{This work is published in IEEE Access.\\DOI: 10.1109/ACCESS.2022.3230590}}

\begin{abstract}

Scene graph generation from images is a task of great
interest to applications such as robotics, because graphs are the main way to
represent knowledge about the world and regulate human-robot interactions in
tasks such as Visual Question Answering (VQA).
Unfortunately, its corresponding area of machine learning is still relatively in
its infancy, and the solutions currently offered do not specialize well in
concrete usage scenarios.
Specifically, they do not take existing ``expert'' knowledge about the domain world
into account; and that might indeed be necessary in order to provide the level of
reliability demanded by the use case scenarios.
In this paper, we
propose an initial approximation to a framework called Ontology-Guided Scene Graph
Generation (OG-SGG), that can improve the performance of an existing machine
learning based scene graph generator using prior knowledge
supplied in the form of an ontology (specifically, using the axioms defined within);
and we present results evaluated on
a specific scenario founded in telepresence robotics.
These results show quantitative and qualitative improvements in the
generated scene graphs.

\end{abstract}

\section{Introduction}

\begin{figure}[t]
	\centering
	\insetfont

	\includegraphics[width=.8\columnwidth]{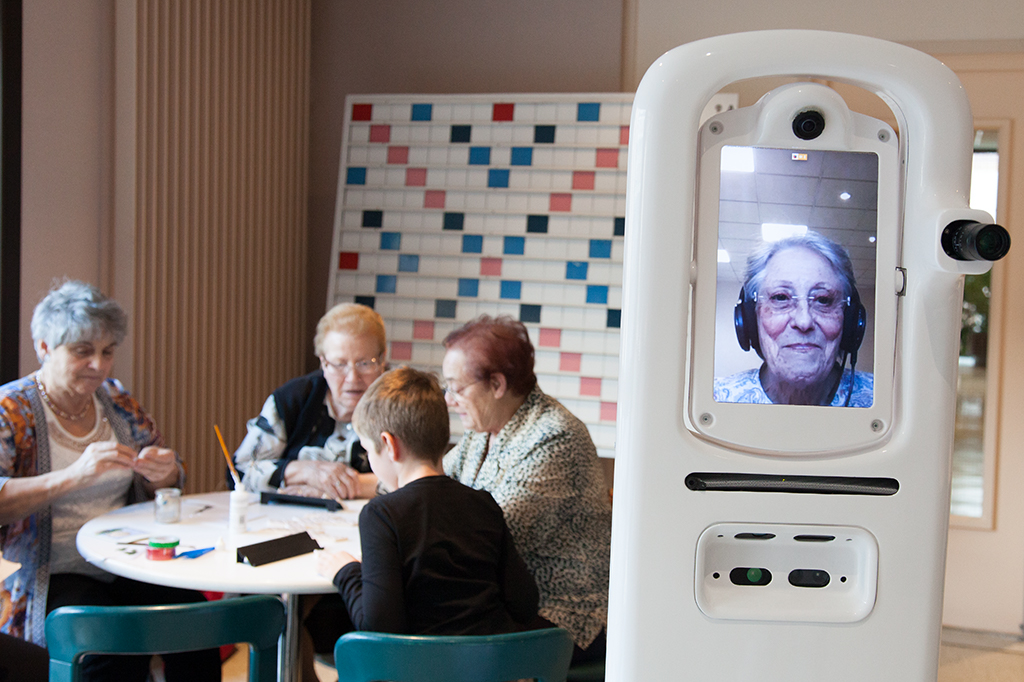}

	\caption{A telepresence robot for elderly care. The robot in this picture is
	used to remotely assist with the activities of a day care centre.}

	\label{fig:teresa}
\end{figure}

Telepresence robots allow people to remotely interact with others.
They are often called ``Skype on a stick'' because they combine the conversation
capabilities of teleconference software with the mobility of robots controlled by
humans for a better social interaction \cite{Kristoffersson:2013:RMR}.
They are also sometimes referred to as ``your alter-ego on wheels'' because they
have a clear application in assistance tasks. For example,
they allow disabled people to attend events remotely,
or caregivers to interact remotely with people under their care.
In particular, this work considers the application of telepresence robots
for elderly care \cite{teresa_icra2015} (see Fig.~\ref{fig:teresa}).

However, controlling these systems is a complex task.
The human user needs to focus on both low-level tasks (such as controlling the robot)
and high-level tasks (such as maintaining a conversation) at the same time; and
this can lead to a cognitive overload, therefore reducing the attention that is
given to the high-level tasks \cite{tsui2011exploring}.
One approach to reduce this overload involves leveraging semi-autonomous capabilities
to allow the user to control the robot using only high-level commands
(i.e., \textit{Approach a given object}, \textit{Follow a given person}, etc),
while the robot takes care of low-level control.
This is in fact a necessity if one considers visually-impaired people as users
of the robotic system.

In all these cases, the robot needs to extract and provide semantic information
about the scenario so that the scene can be described in human terms to the users,
and they can in turn indicate the robot where to go next for interactions.
At the same time, we need a representation of this information that allows
the robot to perform automated reasoning.

This work considers recent advances in visual scene graph generation
(i.e. \cite{WANG202155,9157682,koner2020relation,zellers2018scenegraphs})
and investigate their application to telepresence robots.
These methods extract a semantic graph for a given image, composed of the
main objects present in the scene and relations between them.
The main problem with current solutions is that they are aimed at general scenarios
and do not take existing ``expert'' knowledge about the domain world
into account; and that might indeed be necessary in order to provide the level
of reliability demanded by the usage scenarios.
For example, it is easy and intuitive for a human to understand that a
\texttt{Person} can only be sitting on a single \texttt{Chair} at a given time.
These current solutions have no means of providing common-sense axioms that
can help filter out inconsistent detections, and thus they can erroneously
predict that a person is sitting on multiple chairs at the same time.
These axioms can be collected using what is known as an \textit{ontology},
and it is the main pillar that enables this work.

The main goal of this work is thus finding a way to reuse and repurpose existing scene
graph generation models and datasets for specific robotic applications, and
applying additional techniques that take into account existing domain knowledge of the
application, so that we can improve the performance of a machine learning model
within the reduced scope of a given problem and ontology.
In particular, this work proposes the following contributions:

\begin{itemize}
	\item A methodology for augmenting an existing scene graph generator with
	the addition of steps involving ontology-founded reasoning,
	as opposed to simply defining a new model.
	\item A recipe to prepare an existing dataset for a desired application by
	applying an ontology, filtering irrelevant information and including
	inferred knowledge as a form of data augmentation.
\end{itemize}

The approach is validated in a real transfer learning case for ontology-based
generation, considering a small specialized dataset and ontology founded in a
robotics application.

In Section~\ref{sec:bg}, we explain the theoretical background surrounding ontologies and
scene graph generation.
In Section~\ref{sec:relw}, we survey the state of the art in several directions, including
methodologies and scene graph generation datasets and models.
In Section~\ref{sec:mthd}, we introduce and explain the components of OG-SGG.
In Section~\ref{sec:expsetup}, we detail the complete setup used to perform experiments,
including the chosen scene graph generation model, as well as the metrics used.
In Section~\ref{sec:teresa}, we show the results of our first experiment using data collected
with our own robot.
In Section~\ref{sec:ai2thor}, we apply the same experiment to an existing robotics scenario.
Finally,
in Section~\ref{sec:othermodel}, we perform additional experiments applying OG-SGG to
other scene graph generation models.

\section{Background} \label{sec:bg}

\begin{figure*}[t!]
	\centering
	\insetfont

	\includegraphics[width=.4\linewidth]{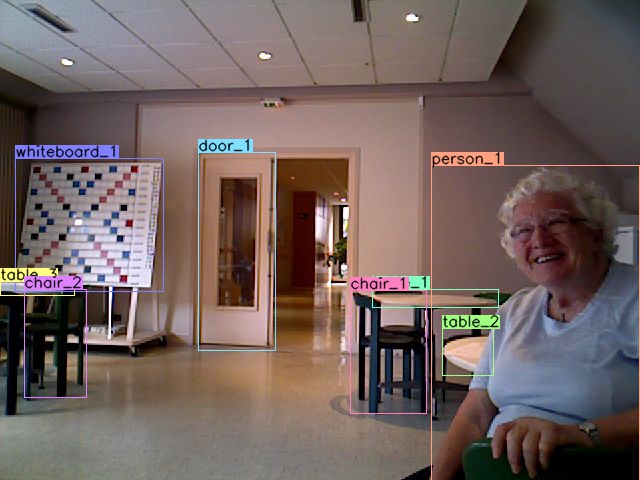}
	~
	\includegraphics[width=.35\linewidth]{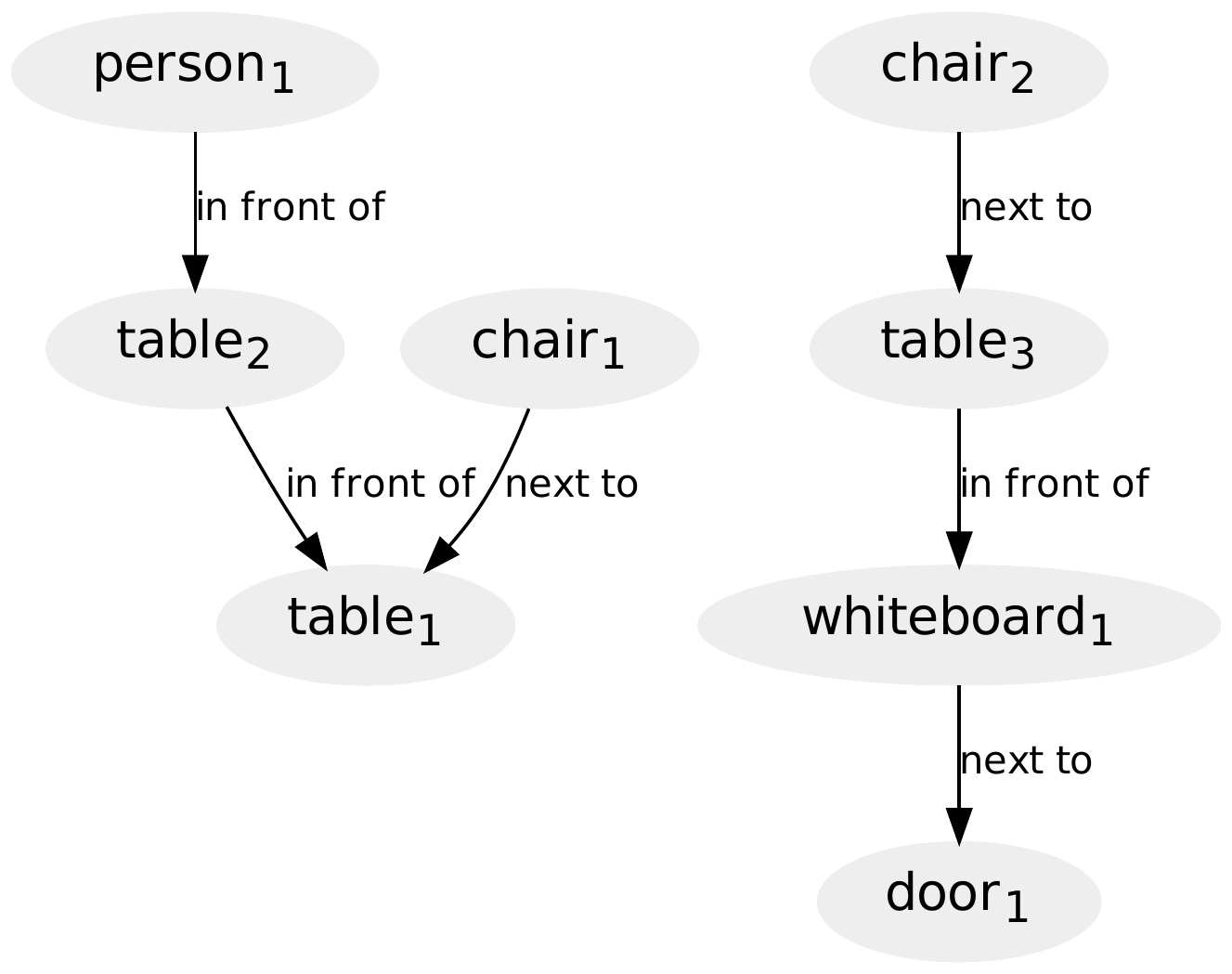}

	\caption{An image captured by the robot, manually annotated with a scene graph, part of the TERESA dataset.}
	\label{fig:examplegraph}
\end{figure*}

\subsection{Ontologies and scene graphs}

Ontologies are broad constructs that can be used to represent the
cognitive model of a given domain world \cite{wiao}. In simplified terms, an
ontology defines a class hierarchy of objects that can exist in the
world, as well as the different types of relations between the objects of the world
(called predicates). Most importantly, an ontology is able to define axioms that
restrict how the predicates can be applied, in addition to producing implicit,
reasoned knowledge from a set of assertions made within the scope of the ontology.
For this reason they are the tool of
choice to represent knowledge bases in robotics and other fields
\cite{7084073,olivares2019review} and perform context-awareness reasoning
\cite{armand2014ontology, diaz2016semantic}.
Ontologies can also be used to model knowledge graphs with richer and more
formal semantics, allowing for higher order reasoning.

This work is based on OWL~2 \cite{w3c2012owl2}, the standard knowledge
representation language for defining ontologies created by the World Wide Web Consortium (W3C).
OWL is built upon RDF \cite{karvinen2019rdf}, which is an earlier W3C XML
standard with the purpose of facilitating data interchange on the Web.
Ontologies are a suitable tool to achieve explainable ML models \cite{arrieta2020explainable}
in the form of knowledge graphs and other semantic web technologies \cite{seeliger2019semantic}.

Given a scene and an ontology,
it is possible to build a scene graph by defining a set of objects
${O = \{o_1, o_2, \ldots, o_n\}}$ (from the classes defined in the ontology) that appear within it, along with a
set $R$ of asserted relation triplets ${(o_i,p,o_j)}$, where
$o_i$ is the source object of the relation,
$o_j$ is the destination object,
${i \neq j}$,
and $p$ is the predicate that describes the relation.
In addition, we can define
${P_{i j} = \{ p_k \mid (o_i,p_k,o_j) \in R \}}$,
which is the set of
predicates for which a corresponding relation triplet exists along the
object pair ${(o_i,o_j)}$. Since there are $n$ objects in the scene,
we can conclude that there are ${n(n-1)}$ object pairs, each with an
associated $P_{i j}$ predicate set.

\subsection{Scene graph generation}

A scene graph generator is a system that, given an input corresponding to a
particular scene (most often an image together with object detection information),
predicts the contents of the $P_{i j}$ predicate set for all given
object pairs ${(o_i,o_j)}$. There are several ways to implement a scene graph generator,
including classical methods based on hardcoded rules, but the most promising
area of research nowadays involves neural networks based on supervised deep
learning. This is the approach considered in this work.

Scene graph datasets are collections of annotated scenes (images) intended
for evaluating scene graph generators, as well as training the aforementioned
scene graph generation networks.
Each image in the dataset is annotated with its associated set of objects $O$
and set of relation triplets $R$. In particular, $O$ is usually
annotated as a series of bounding boxes with class information, each
corresponding to an object;
whereas $R$ is annotated as a list of ${(o_i,o_j)}$ object pairs with
their corresponding $P_{i j}$ predicate set.

A common approach to building scene graph generation networks
involves predicting a ranking score for all possible relation
triplets across the entire image, where higher ranking triplets
are deemed as more likely to occur than lower ranking triplets.
A trimming operation takes place afterwards, which removes triplets
according to a given set of criteria. The conventional way to define this
operation is Top-K, which results in the K highest ranking triplets
being retained and the rest being discarded.

This work proposes a series of techniques to improve this pipeline
by involving axioms defined in the ontology that govern the
semantics of the predicates. These axioms are used to augment the
source data used during training and control the trimming operation
that affects the network's output. Specifically, the following types of
axioms that affect predicates, which are defined by \cite{w3c2012owl2},
have been considered:

\begin{itemize}
	\item Domain and range restrictions: these axioms assert that only
	objects belonging to certain classes can be the source or destination
	of a predicate (respectively).
	For example, we can say that for the predicate \texttt{sitting on},
	the domain is \texttt{Person} and the range \texttt{Chair} --
	it is not possible to say
	\texttt{(plant\textsubscript{1},sitting on,food\textsubscript{1})}.
	\item Inverse relationships: these axioms assert that one predicate
	is the inverse of another (with the source and destination objects
	inverted).
	For example, we can say that the predicates \texttt{on top of}
	and \texttt{below} are inverses of one another.
	\item Transitivity: these axioms assert that, if two relation triplets
	${(o_i,p,o_j)}$ and ${(o_j,p,o_k)}$ are given, then ${(o_i,p,o_k)}$ also holds.
	For example, we can say that the predicate \texttt{behind} is transitive,
	since if both
	\texttt{(person\textsubscript{1}, behind, chair\textsubscript{1})} and
	\texttt{(chair\textsubscript{1}, behind, table\textsubscript{1})} hold,
	then \texttt{(person\textsubscript{1}, behind, table\textsubscript{1})} must also hold.
	\item Functionality: these axioms assert that there can only be one
	object related by a predicate to a different one. For example,
	the predicate \texttt{holding} can only accept one source object
	for each destination object --
	if \texttt{(person\textsubscript{1},holding,pencil\textsubscript{1})} holds,
	then no other \texttt{Person} object can be related to
	\texttt{pencil\textsubscript{1}} through \texttt{holding}.
	\item Symmetry: these axioms assert that a certain predicate does not
	mandate an order in which the two objects are related. This means that
	if the source object is related to the destination object through the predicate,
	then the destination object is also related to the source object through the
	same predicate. For example,
	\texttt{(chair\textsubscript{1},next to,table\textsubscript{1})} implies
	\texttt{(table\textsubscript{1},next to,chair\textsubscript{1})}.
\end{itemize}

\section{Related work} \label{sec:relw}

Given the previously mentioned use case in telepresence robotics, we initially surveyed simpler,
more direct approaches such as automatic image captioning \cite{sat2015},
combined with refinement based on data sampled from our own robot. The main
problem with this approach had to do with the lack of structure and lack of
coherence in generated captions (which was confirmed by other researchers using indicators
such as Semantic Fidelity \cite{agarwal2020egoshots}), resulting in
unsatisfactory results from the point of view of potential users, as well as
lack of usability for downstream robotic tasks.

We quickly learned that a more formal and richer way of representing knowledge
about environments was necessary. This led us to shift our attention towards scene graphs,
specifically existing work concerning their automatic generation.
This section contains a detailed survey of the state of the art, as well as what
we believe to be their relevance and contribution to solving domain specific
problems like ours.

\subsection{Methodologies}

A line of research that we deemed relevant revolves around neurosymbolic
(NeSy) computation approaches. This interest arose from our need to improve the
specialized application of a deep learning model towards a certain application.
Several approaches exist, including ones that are oriented
towards improving the explainability of AI solutions (XAI) \cite{diaz2022explainable, bennetot2019towards},
an increasingly important topic.

Some approaches such as \cite{donadello2019compensating} propose applying NeSy
logic constraints in the form of Logic Tensor Networks \cite{badreddine2022logic}
to scene graph generation. These approaches map discrete logical operators to
continuous differentiable operators based around fuzzy logic. However, they only
consider the traditional approach of using training and testing splits of the
same dataset, leaving out the desired goal of transferring knowledge learned from
one dataset to a different scenario. Moreover, a new model architecture specifically
designed to incorporate NeSy components had to be created, which might not always be
feasible to implement depending on the constraints of the intended application.

\subsection{Scene graph datasets}

The main ingredient that makes or breaks a machine learning application
is the dataset. A good dataset, although not a guarantee of success,
is a necessary precondition. Currently there exist several of them which
have some relevance to the domain-specific scene graph generation task:

\begin{itemize}
	\item MS~COCO \cite{10.1007/978-3-319-10602-1_48} is the de-facto
	standard dataset for image classification, segmentation, captioning and
	object detection. However, despite offering 5 free-form textual captions
	per image, it does not contain any usable semantic information
	needed to make a machine learn how to generate scene graphs.
	Nonetheless, other researchers have built upon MS~COCO in order to
	create scene graph datasets, such as those which will be discussed
	next.
	\item Semantic PASCAL-Part \cite{semanticpascalpart2016} is an
	OWL conversion of an earlier PASCAL-Part dataset \cite{Chen2014DetectWY}.
	Formal ontological constructs are used to define each object class,
	which makes this dataset appropriate for object classification tasks
	based on detection of constituent parts. Unfortunately, this also means that
	the dataset is not suitable either for downstream scene graph tasks,
	as there is effectively only one possible ``relation'':
	\texttt{isPartOf} (and its inverse \texttt{hasParts}).
	\item VRD \cite{lu2016visual} (Visual Relationship Detection) is
	one of the first datasets developed during scene graph generation
	research. It is a relabelling of an earlier Scene Graph dataset
	\cite{7298990}, which was itself sampled from the intersection
	of MS~COCO and YFCC100m \cite{10.1145/2812802}.
	\item VG \cite{krishna2017visual} (Visual Genome)\footnote{\url{http://visualgenome.org/}} is a follow-up
	work to VRD that opens up the annotations to cover the entire
	intersection between MS~COCO and YFCC100m instead of a hand picked
	sample. It was created by crowd sourcing, and it brings additional
	ground truths such as relations between the objects or visual
	question answering examples.
	It makes use of WordNet \cite{10.1145/219717.219748} to
	identify objects and relations,
	which adds a considerable depth to the labelling compared to MS~COCO
	with its 80 broad categories. However, since
	the semantic data is generated automatically from the crowd
	sourced input, it is quite noisy and thus requires serious
	preprocessing before it can be used. The maintainers of VG also offer
	a list of duplicate/aliased object and relation classes that is
	nearly always used as the first step of the required preprocessing.
	\item VG-SGG is a preprocessed version of VG introduced by
	\cite{8099813} which has subsequently been adopted by researchers as
	the VG split of choice for training and evaluating scene graph generation
	networks, hence the name. Bounding box information is cleaned up, and
	only the 150 most frequent object classes and 50 predicate classes are
	used.
	\item VrR-VG \cite{liang2019vrrvg} (Visual-relevance Relations) is
	another filtered and improved version of VG specifically intended for scene
	graph generation. It improves VG by removing from the dataset
	high frequency, low quality ambiguous relations that can be easily
	detected with mere probabilistic analysis;
	and leaving smaller, high quality
	ones that require visual and semantic reasoning to detect.
	\item GQA \cite{hudson2018gqa} (Graph Question Answering) is yet another
	dataset based on VG, but focused on visual question answering. Even though
	it is intended to be used to solve a different task it is still
	of interest, because it contains scene graphs with object information
	that has been further cleaned, filtered and even manually validated.
	In addition, it augments images with a \texttt{location} annotation
	that discloses the type of environment (indoors or outdoors).
\end{itemize}

Overall, the scope of existing datasets tends to be very broad,
aiming to fulfill the general use case (without any specific domain in mind).
In addition, they exhibit great imbalance of object and relation classes,
which makes it more difficult to train domain specific models, as well as
resulting in undesirable bias towards the few most overwhelmingly common classes.
Furthermore, the labels are usually noisy and sparse (incomplete), which results
in a risk of the network learning fictitious unintended patterns.
Finally, none of the existing datasets have a formally defined ontology,
instead relying on free form annotation of objects and relations devoid
of any specific semantics that could be used to extract additional inferred
knowledge, or improve the quality of the generated graphs.

This work proposes methods by which existing scene graph datasets
can be adapted to fit within an existing ontology, as well as enriching
the annotations through inferred knowledge derived from the axioms in the
ontology.
The desired end goal of this endeavor is making it possible to reuse existing
data to solve new domain-specific scene graph generation problems.

\subsection{Scene graph generation}

Researchers have been iterating over different ideas on how to approach the
scene graph generation problem. Below is a summary of some of the most
interesting approaches that have been published:

\begin{itemize}
	\item The original VRD model \cite{lu2016visual} introduced alongside the
	dataset proposed a simple network based on two modules that looked at
	visual and language features respectively, and incorporated likelihood
	priors based on the predicate frequency distribution for a given pair
	of object classes.
	\item Iterative Message Passing \cite{8099813} proposed using RNNs to
	iteratively ``pass'' information between proposed edges of the graph in
	order to further refine them using their neighboring context.
	\item Neural Motifs \cite{zellers2018scenegraphs} introduced a new
	architecture based on bidirectional LSTMs that was capable of detecting
	patterns in the structure of the scene graphs called ``motifs.''
	\item VRD-DSR \cite{liang2018Visual} proposed combining visual appearance,
	spatial location and semantic embedding ``cues'' in a single network, as
	well as treating scene graph generation as a triplet ranking problem.
	\item Unbiased Causal TDE \cite{9157682} proposed a new scene graph
	benchmark framework with better defined metrics, along with a new
	model agnostic technique that aims to reduce bias during training.
	\item Schemata \cite{sharifzadeh2020classification} were proposed as a way
	of introducing an inductive bias in the form of relational encoding that
	allows the network to learn better representations from the training data
	as non-expert prior knowledge, resulting in better generalization. This encoding
	can also be propagated during model fine-tuning with additional external
	triplet data, without the need for image data.
	\item VRD-RANS \cite{WANG202155} improved upon VRD-DSR by changing the
	visual feature extraction, adding a recursive attention module with a GRU,
	and integrating a form of data augmentation based on negative sampling into the
	training pipeline.
	\item RTN \cite{koner2020relation} applies the Transformer architecture
	to scene graph generation for the first time. It follows a conservative
	approach in regards to the inputs used and its usage of a likelihood prior,
	and introduces the concept of positional embedding for nodes and edges.
\end{itemize}

In general, analogously to the corresponding work on creating datasets,
these models aim to solve the general problem of scene graph generation.
Most if not all models are based around a traditional two-stage object detector
such as Faster-RCNN \cite{NIPS2015_14bfa6bb} (close, but not quite usable in realtime),
and many make use of the intermediate feature maps extracted for specific
regions of interest prior to the object classification stage.
In addition, existing codebases are geared towards training and evaluating
the models (including the object detectors) on existing datasets,
with little to no thought put into transfer learning tasks, custom datasets,
nor evaluations of individual components. These factors make it needlessly
difficult to adapt existing scene graph generation solutions to new problem
domains such as the previously mentioned robotics application.

Another problem we identified with existing solutions is that they are designed
as pure deep learning architectures and, as such, they cannot take advantage of subtle
semantics concerning the defined object and relation classes that are implicit and
intuitive for humans. Thus, it is common for SGG models to output scene graphs
full of inconsistencies. The methodology
proposed in this work is capable of improving the usability of scene
graph generation for specific domains, and is independent of the model used.
That is, it can be adapted for use with any specific scene graph generation
model that might be best suited within the constraints imposed by the problem
(such as efficiency, for example).

\subsection{Robotics related research}

Several works with some relevance to this problem have previously been published,
including the following:

\begin{itemize}
	\item Ontologenius \cite{8956305} is a semantic memory module
	based on OWL ontologies for the ROS \cite{ros2009} environment. It can
	be used to store the knowledge of the robotic agent, as well as to perform
	reasoning on it. However, it does not contain any perception
	functionality, meaning it needs to be supplied externally
	with knowledge by other nodes within the ROS environment. Some research
	papers such as \cite{9223485} make use of it as the underlying knowledge engine.
	\item RoboSherlock \cite{BalintBenczedi2019RoboSherlockCR} is a framework
	for cognitive perception based on unstructured information management.
	It offers perception related functionality, but the implementation is
	based on classical algorithms instead of deep neural networks, which makes
	it difficult to generalize to new environments, situations or use cases.
	\item Other research such as \cite{robotfailures2021,semanticrobotfailures2021} makes use of
	deep learning models to construct scene graphs in robotic contexts,
	but their approach is also limited in scope and application.
	In addition, the potential for using the internal structure of the
	ontology to guide the process is left untapped; instead still relying on
	labels devoid of any semantics, i.e. without a formally defined ontology
	that describes the class hierarchy and axioms governing the predicates.
\end{itemize}

Overall, there are interesting building blocks that can be used as reference
for developing new applications. However, the usage of ontologies for semantic
perception is still fairly young.
Moreover, researchers prefer placing their focus on solving
specific problems by sacrificing the potential for generality.
Even though this work showcases a specific application, it aims to provide
methods that can be reused in other applications without significantly
changing how they work.

\section{Methods} \label{sec:mthd}

\begin{figure*}[htbp!]
	\centering
	\insetfont

	\includegraphics[width=0.9\linewidth]{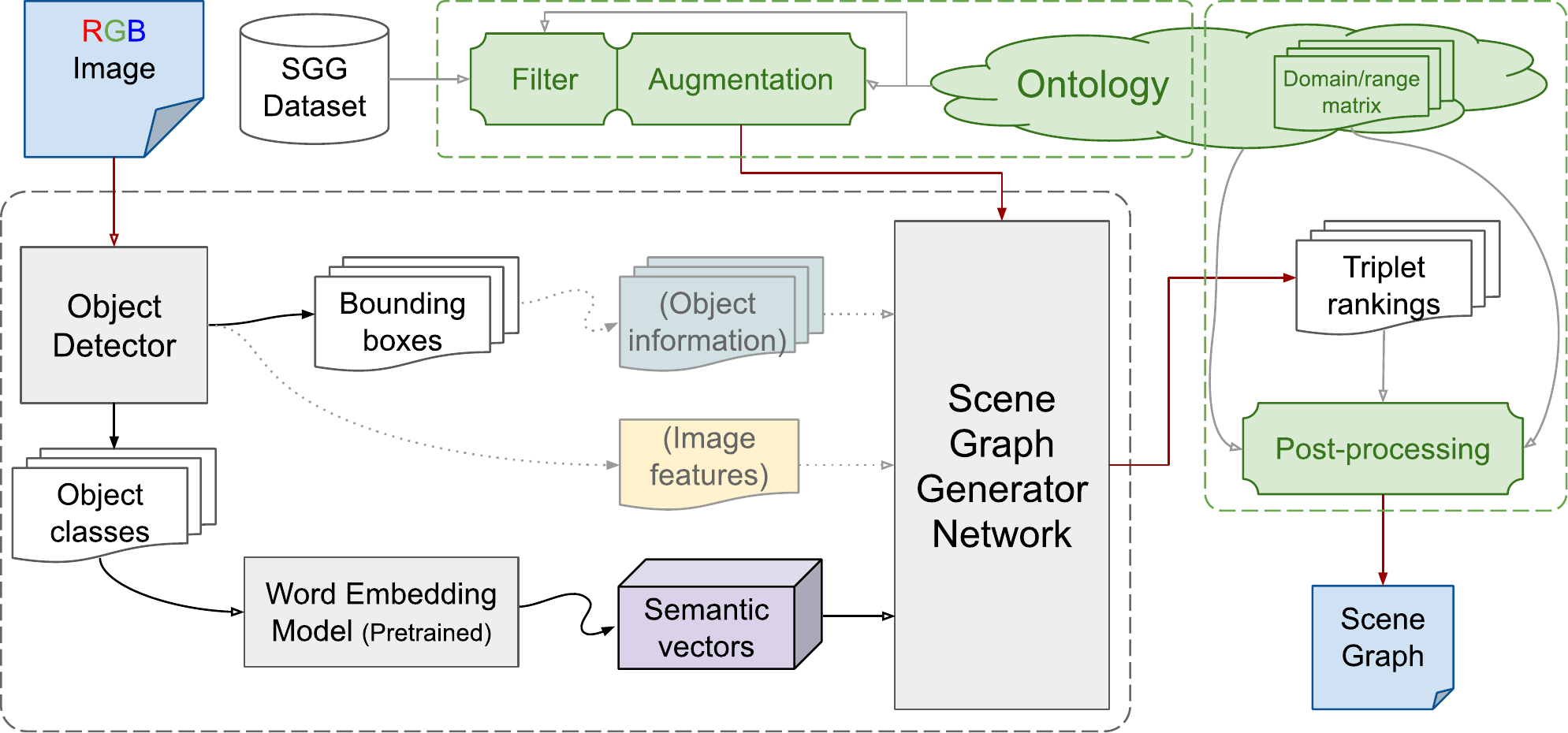}

	\caption{Full OG-SGG pipeline proposed protocol. The diagram shows its three main components
	in full detail, along with both internal and external data flows.
	The proposed ontology-aware additions are highlighted in green,
	which include the filter/data augmentation component, and the
	post-processing component.
	A pre-existing object detection and scene graph generation component
	is also included in the pipeline. The latter receives semantic vectors
	as input, and optionally other data such as object location information
	and visual feature maps.}

	\label{fig:pipeline}
\end{figure*}

The proposed pipeline, illustrated in Figure~\ref{fig:pipeline},
consists of three main components: a scene graph generation
network, a training dataset filtering and augmentation process, and a network
output post-processing process. These last two processes, the core of this work's
contribution, make use of pre-existing expert knowledge defined in the domain
ontology, whereas the network itself can be adapted from the existing state
of the art with minimal changes according to needs (such as efficiency).

\subsection{Scene graph generation network}

OG-SGG augments an existing scene graph generation network. The only requirements
imposed by our proposed methodology on this SGG network are the following:

\begin{itemize}
	\item It needs to be able to detect and process objects within the image
	(bounding box predictions being sufficient, although other approaches
	such as image segmentation could potentially provide more precision in
	the localization of the objects, especially in crowded scenes).
	The object detection can be done either as
	part of a combined object detection--relationship detection network, or
	as a standalone component that reuses the output of an existing object detector.
	In later sections of this paper, we focus on the latter case in order to evaluate our methodology purely
	on the quality of the detected relationships, and not the object detection.
	\item It needs to receive semantic information about the types of objects
	within the scene in the form of dense semantic vectors, as opposed to other
	ways such as one-hot encodings. Each object class is assigned a different
	semantic vector, in turn sourced from an existing corpus of pre-trained
	word embedding vectors. This is intended to enable the generalization
	capability of the network, and is in fact required so that the network
	can be repurposed for different sets of possible input objects without
	needing retraining.
	\item It needs to output a single ranking value for every possible
	knowledge triplet proposal, across all detected objects and proposed
	relationship types -- no range restrictions are applied.
\end{itemize}

Once all triplet ranking scores are calculated, they are sorted in descending order. The
threshold below which triplet proposals are judged as more unlikely than
likely is usually left undefined, therefore requiring users of the network to
establish their own post-processing rules. A later section of this work
will revisit this area, where post-processing rules that take
into account information expressed in the ontology will be proposed.

\subsection{Dataset filtering and data augmentation}

We decided to focus this work on reusing existing scene graph datasets and repurposing
them to be usable within the scope of our problem, which consists in describing
a scene with a knowledge graph. For this, we
defined a formal ontology for the target problem, and applied a series
of ontology-guided transformations to the source dataset.

First of all, we parse the original source format of the input data
and convert it to a common representation.
During this step, an initial form of filtering based
on ad-hoc constraints can be carried out (such as, for example,
removing all images not tagged in a particular way,
e.g. ``indoor'' images; or not containing
specific objects, and so on). This has the (possibly desired) side
effect of reducing the size of the dataset while maximizing or maintaining
its quality in terms of performance.
In a later section we
will show how the filtering can sometimes even lead to improvements
in the results.

Next, it is necessary to convert object class annotations into semantic
vectors that can be fed to the network.
We decided to use an existing word embedding model
pre-trained on the
English Wikipedia corpus\footnote{\url{https://tfhub.dev/google/Wiki-words-250-with-normalization/2}}
in order to generate the semantic vectors.
Mapping objects from the source set of classes to the ontology's set
of classes was deemed unnecessary, given the generalization
power of training the network on a much richer set of semantic vectors than
the one that can be derived from the reduced set of classes in the ontology.

\begin{figure}[htbp!]
	\centering
	\insetfont

	\includegraphics[width=0.9\linewidth]{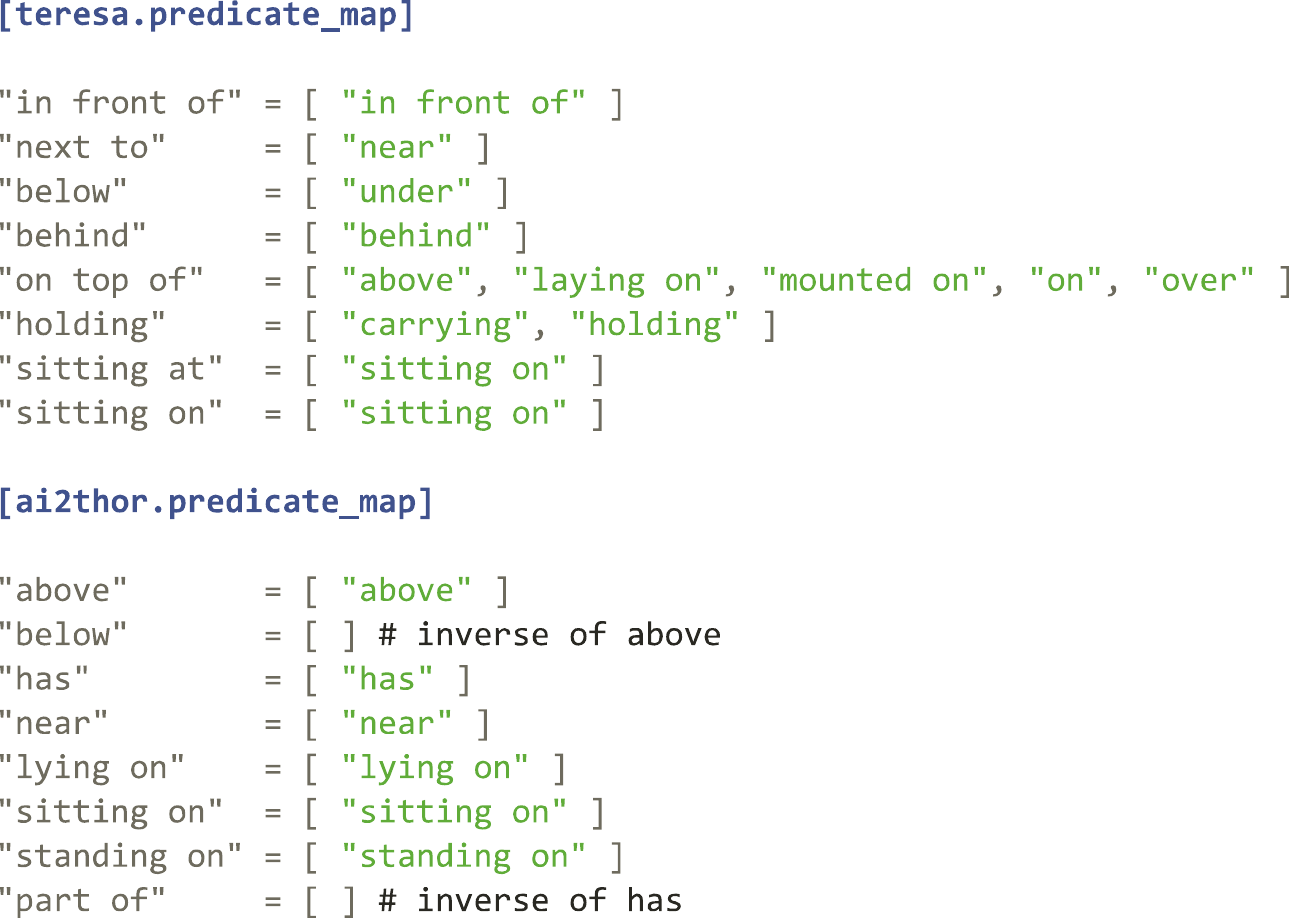}

	\caption{Examples of predicate maps. Each key corresponds to a predicate
	in the ontology, whereas each value is a list of equivalent predicates
	in the source dataset.}

	\label{fig:predmap}
\end{figure}

Following that, predicates defined by the source scene graph dataset need to be
mapped into the corresponding predicates of interest defined by the given problem
domain's ontology (see Fig.~\ref{fig:predmap}).
In order to do this, we manually defined the correspondence between the two sets
of predicates, which is then used during this process to translate the predicate
component of each relation triplet.
Additionally, we
discarded relation triplets that contain predicates not matched
with any in the ontology.
This mapping is the only information that needs to be externally defined
and provided during the process, besides the ontology itself.

Once all triplets are using predicates defined in the ontology,
we feed each scene in the dataset to
an OWL processor module previously initialized with the ontology.
We selected Owlready2 \cite{LAMY201711}
as the software library providing ontology processing.
This ontology processor is able to load and parse ontologies in the OWL format,
perform inferences on the provided knowledge using the axioms defined in the
ontology, and generate implicit triplets.
This includes the generation of triplets for inverse, symmetric and transitive predicates.
For example, \texttt{(chair\textsubscript{1}, next to, table\textsubscript{1})}
would generate \texttt{(table\textsubscript{1}, next to, chair\textsubscript{1})}.
Likewise, \texttt{(cup\textsubscript{1}, on top of, table\textsubscript{1})}
would generate \texttt{(table\textsubscript{1}, below, cup\textsubscript{1})}.
These new
triplets are then extracted from the ontology processor and added
back to the dataset, thus resulting in a form of data augmentation.

Finally, once the enriched information obtained through ontological inferences is
extracted from the ontology processor, the final combined data can be divided into
training and validation splits.
Note that the test split of the original dataset is never used, as we are only
interested in converting training data.
We set up this arrangement in order to tune the hyperparameters of the model
and implement early stopping in the training process.
Stratification is applied in order to preserve the frequency distribution of
predicates, which needs to be as similar (and complete) as possible between
the two splits.
In order to stratify a multi-label multi-class data structure such as scene
graphs, we decided to first tally how many images a given predicate appears in,
and then assign the images into buckets corresponding
to the least frequent predicate classes that appear in each.
Afterwards,
the bucket for the overall least frequent predicate can be selected
and its images added to the stratified splits according
to the desired proportion.
Since we removed images from circulation by doing this, the frequency
distribution must be recalculated and the bucket assignment redone.
We repeat this process until all predicates are processed.

\subsection{Output postprocessing}

\begin{figure}[htbp!]
	\centering
	\insetfont

	\includegraphics[width=\linewidth]{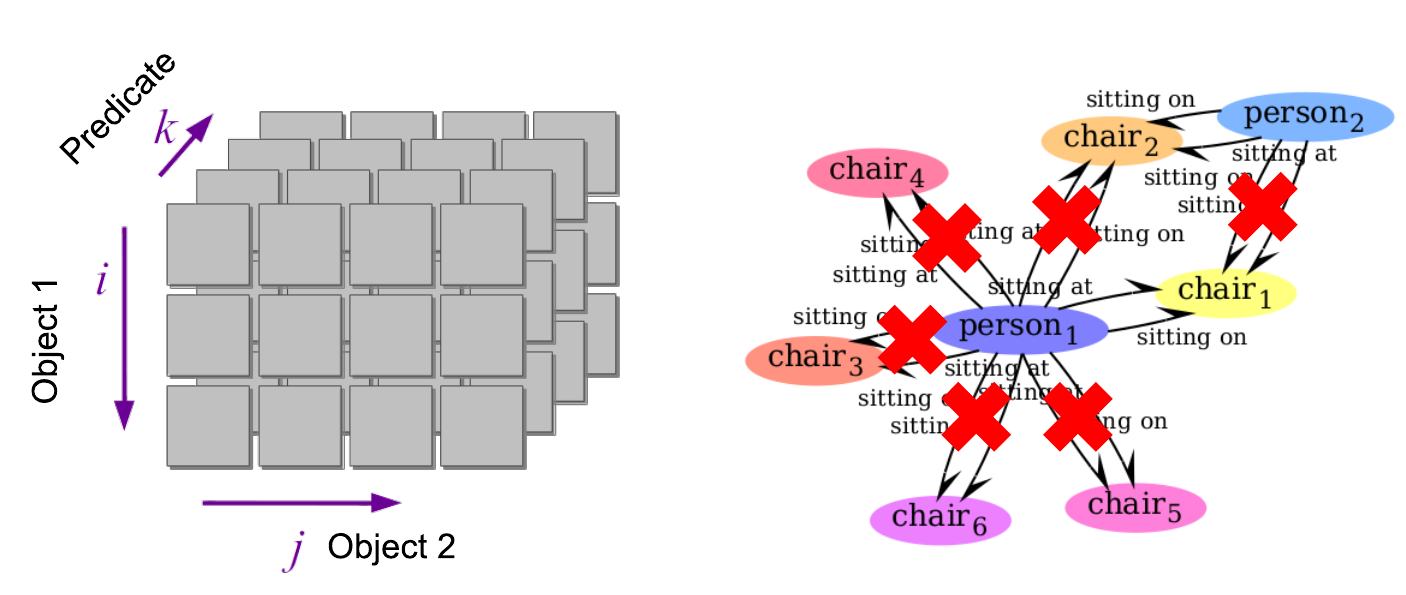}

	\caption{Different post-processing techniques proposed in this section.
	\textbf{Left}: Domain/range constraint tensor $C$. \textbf{Right}: Axiom-based pruning.}

	\label{fig:postproc}
\end{figure}

The axioms defined in the ontology restrict the set of possible
relation triplets that can appear in a scene, and thus can be used
to filter out predictions that we know beforehand to be invalid.

We propose appending a new output post-processing stage to the prediction process
that prunes relation triplets that introduce violations of axioms defined in
the ontology (see Fig.~\ref{fig:postproc}). We studied in Section~\ref{sec:bg} the leveraging of several kinds of axioms that affect
predicates, such as Functional/InverseFunctional restrictions, or domain/range
restrictions. The general filtering approach consists of only accepting the
highest ranking mutually exclusive triplet proposals and pruning the rest.
For example, if there are two triplet proposals,
\texttt{(person\textsubscript{1}, sitting on, chair\textsubscript{1})} with score
$0.78$ and \texttt{(person\textsubscript{1}, sitting on, chair\textsubscript{2})}
with score $-0.4$, the first triplet is accepted and the second one is pruned.

Additionally,
we decided to study domain/range axioms, in particular using the
internal semantic structure of the relationships between object classes in the ontology.
We express the domain/range axioms as a boolean tensor $C$ with the
shape $|O| \times |O| \times |P|$ (see Fig.~\ref{fig:postproc}),
where the first two dimensions correspond to the object classes in the ontology
(i.e. a generic pair of objects) and the last dimension to the predicate classes
-- this unusual ordering is used in order to improve the efficiency of the retrieval
of predicate compatibility information according to a given object pair used as key.
An element of the tensor is True if its
associated predicate is compatible with the domain and range
corresponding to the given object classes, and False otherwise.
Predicted triplets from the output of the model can thus be individually looked
up in $C$ and kept or discarded according to the truth value.

We compute $C$ beforehand by programmatically introspecting on the ontology,
matching up all possible pairs of object classes, and verifying
their compatibility with all predicate classes.
We implemented
support in the introspection code for
a basic set of logic constructs found in OWL that
are used to define the domain/range of a predicate. This includes
the \texttt{And}, \texttt{Or} and \texttt{Not} operators;
as well as support for walking through the class hierarchy
(e.g. if the range of a property is \texttt{GrabbableObject},
then it will be legal
to have a \texttt{Cup} in the object position).

\section{Experimental setup} \label{sec:expsetup}

We settled on VRD-RANS \cite{WANG202155} as the baseline scene graph generation
network for experiments in this work, which was implemented as faithfully as possible.
This network was chosen for the following reasons:

\begin{itemize}
	\item It only needs a single global feature map extracted from the image,
	as opposed to other solutions which mandate the use of two-stage
	object detectors based on Regions of Interest (RoIs). This allows for
	using quick, robotics oriented single-stage object detectors such as
	those belonging to the YOLO \cite{yolov1} or SSD \cite{ssd} families.
	\item It receives object localization information in the form of binary
	masks, which could be expanded in future work to contain additional
	information captured by the robot.
	\item It contains a recursive attention module, allowing the network to
	focus on processing the most relevant parts of the image at once.
	\item It uses a novel training strategy that consists of providing the
	network with fixed-size batches, one for each image in the dataset, and
	containing examples of both labelled and unlabelled object pairs
	(referred to as \textit{positive} and \textit{negative} examples, respectively).
	The idea behind this is compensating for the sparse nature of datasets,
	and taking advantage of the large number of unannotated pairs for
	data augmentation and regularization purposes.
\end{itemize}

VRD-RANS, like other scene graph generation networks \cite{8099813,zellers2018scenegraphs,9157682},
operates on an object pair by pair basis (each individual corresponds
to a given object pair), and is in charge of predicting ranking scores for
each of the predicates defined by the scene graph dataset.
Generating a scene graph involves feeding all object pairs to the network
in order to extract the ranking scores, which
are raw unbounded values $\in \mathbb{R}$ with no defined semantics other than
comparison operators (i.e. $<, >, \le, \ge, =, \neq$).

\subsection{Network layers}

\begin{figure*}[t]
	\centering
	\insetfont

	\includegraphics[width=0.9\linewidth]{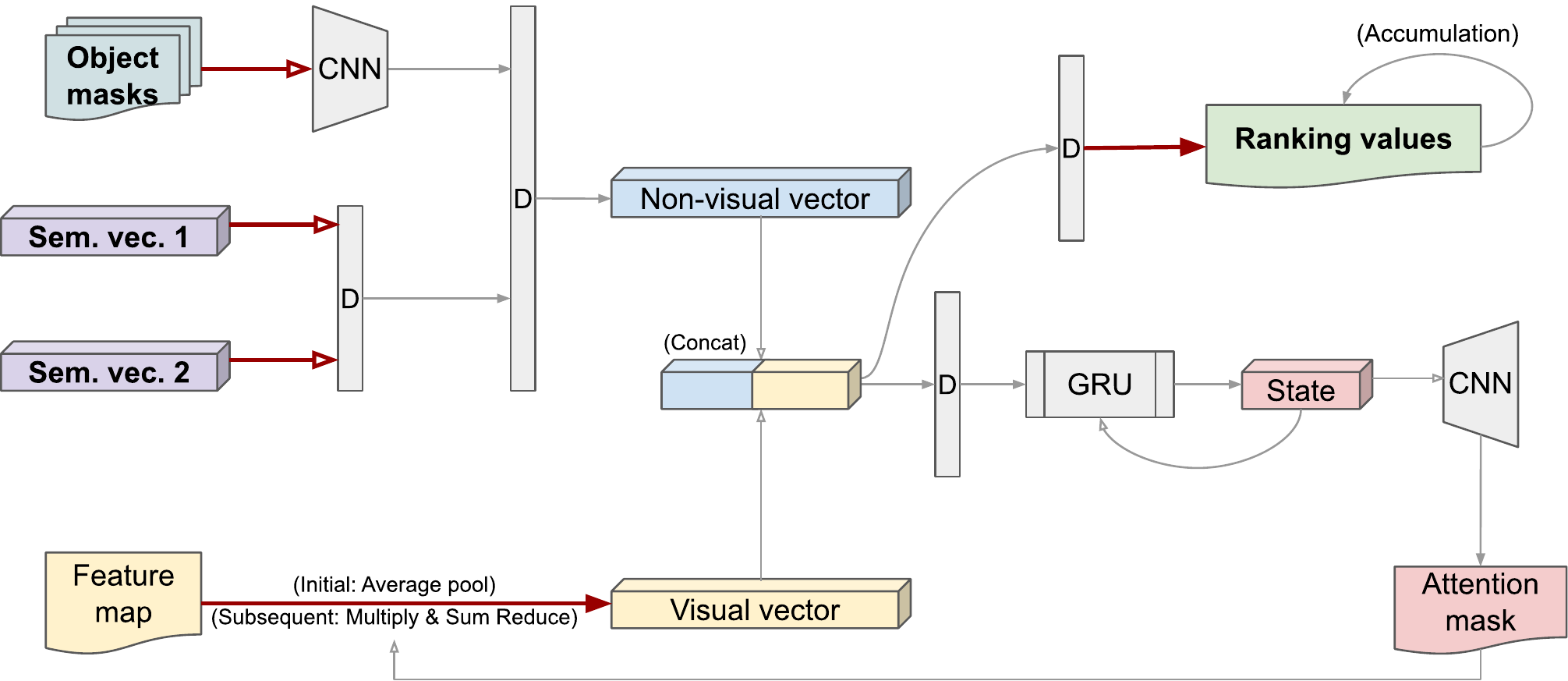}

	\caption{Diagram of the overall network structure of VRD-RANS \cite{WANG202155}.
	This is the chosen scene graph generation network for use in our experiments.
	The network, which is divided into an initial feed-forward part
	and a main recursive part,
	contains several Dense layers (marked with a D);
	two series of CNN layers (one of which is transposed),
	and a single Gated Recurrent Unit (GRU).
	The visual vector is initialized with an average pooling of the
	feature map for the first iteration of the recursive part,
	and in subsequent iterations the averaging is additionally weighted
	using the freshly calculated attention mask.
	The non-visual vector is initialized by the feed-forward part,
	which takes the semantic vectors and object masks as input, and
	remains unmodified throughout the execution of the recursive part.}

	\label{fig:network}
\end{figure*}

The network (illustrated in Figure~\ref{fig:network}) receives four inputs,
three of which are given for each individual
(object pair), and the other one is shared across all individuals belonging to
a given image. Specifically, the network accepts an object mask (two channels,
one for each object in the pair), two semantic vectors, and the global feature
map of the image. Objects masks are simple matrices where each pixel falling
within the bounding box of a detected object is set to 1, and the rest remain 0.
These masks are resized to a fixed dimension in order to improve efficiency.

The first component of the network is the non-visual vector generator.
This module fuses all per-object-pair inputs, and outputs a new vector which is
later used by the core part of the network. This fusing is performed in two steps:
first, the object mask is processed by three convolutional layers followed by a
dense layer, and in parallel the two semantic vectors are concatenated and fed to
two consecutive dense layers. The outputs from these two branches are finally concatenated
in order to form the final non-visual vector.

The core part of the network receives the non-visual vector and the visual feature map
as inputs. Before the main loop, a new \textit{visual vector} variable is initialized by performing
a 2D global average pool of the entire feature map. Additionally an accumulator vector
is initialized with the output of a single dense layer receiving the concatenated visual
and non-visual vectors. This dense layer, which is the final layer of the network, has
as many outputs as predicates defined in the ontology.

The network performs a recursive process with a fixed number of iterations (5 was used, same as \cite{WANG202155}).
In each iteration, the concatenated visual and non-visual vectors are processed by two
other consecutive dense layers before arriving to a GRU. The GRU's hidden state is zero-initialized,
and in each following iteration it will contain the output hidden state from the previous iteration.
The GRU's output is then processed by four chained transposed convolutional layers, which in turn
generate a new attention mask encoding which locations in the image are to be ``looked at'' next.
The size of this attention mask does not match that of the feature map, so it is necessary to resize it
to the same dimensions (using linear interpolation). This new resized attention mask is then used
as the weights for a new 2D global average pool of the feature map (from which a single scalar value
is obtained from each channel). This new output is stored in the visual vector, and finally the
accumulator is updated by adding the new output of the final dense layer to it.

After performing all iterations, the accumulator is divided by one plus the total number of iterations,
calculating thus the arithmetic mean of all outputs provided by the final dense layer. This value is
the final output of the network.

In general, the activation function used throughout the entire network is ReLU, due to its
simplicity and efficiency. In order to improve the stability of the training process,
several batch normalization layers were placed between consecutive dense layers.
Also worth mentioning is the fact that the attention mask is generated using softmax activation
so that it can be used as weights for a weighted mean (in other words, so that all coefficients
add up to 1). The final layer of the network does not use any activation function. This is
required so that ranking values can be generated.

\subsection{Loss function}

As in VRD-RANS \cite{WANG202155}, the loss function used to train the network
is the the multi-label hinge loss margin function.
The following scalar loss value is calculated, cross referencing all object
pairs that appear within the training minibatch:

\begin{equation*}
	\mathcal{L} = \frac{1}{N n} \sum_{\forall i \mid y_{i} = 0} \sum_{\forall j \mid y_{j} = 1 } \max \big( 0, 1 - ( \hat{y}_{j} - \hat{y}_{i} ) \big)
\end{equation*}

\noindent where $N$ is the number of object pairs in the minibatch (i.e.
its size), and $n$ is the number of predicates (i.e. network outputs).
Thus, $Nn$ is the total number of triplet predictions in the minibatch.
$y_i$ is the ground truth value for a given triplet $i$ in the minibatch
(evaluating as 1 if the triplet is present and 0 if not present),
and $\hat{y}_i$ corresponds to the output of the network,
that is, the ranking scores $\in (-\infty, +\infty)$ predicted by
the network for each triplet.
This function, which takes the entire mini-batch output of the network at once,
is thus designed to cause the network to incur a loss when the scores $\hat{y}_{i}$ corresponding
to triplets not present in the ground truth ($i \mid y_{i} = 0$) are ranked higher than those which
are present ($j \mid y_{j} = 1$).

\subsection{Implementation details}

\begin{table}[htbp!]
	\centering
	\insetfont
	\caption{Model hyperparameters used for VRD-RANS}
	\begin{tabular}{ |c|c| }
		\hline
		Hyperparameter & Value \\
		\hline
		Feature map size & $10 \times 10 \times 512$ \\
		Object mask size & $64 \times 64 \times 2$ \\
		Semantic vector size & $250$ \\
		\hline
		Optimizer & AdamW \\
		Learning rate & $10^{-5}$ \\
		Weight decay & $10^{-5}$ \\
		Early stopping patience & $2$ epochs \\
		Batch size & $32$ \\
		\hline
	\end{tabular}
	\label{table:hyparam}
\end{table}

We selected YOLOv4 \cite{bochkovskiy2020yolov4} as the object detection network of
choice given its high real-time performance, suitable for robotics applications.
We used existing weights pretrained on MS~COCO \cite{10.1007/978-3-319-10602-1_48},
which include a CSPDarknet53 \cite{wang2020cspnet} backbone pretrained on
ImageNet \cite{deng2009imagenet}.

The original VRD-RANS \cite{WANG202155} authors seemingly did not publish their code.
For this reason, this work includes a new implementation of their proposed network.
This new implementation was carried out using the TensorFlow framework, with the
high-level Keras API layered on top. A new subclass of \texttt{tf.keras.Model},
called \texttt{RelationshipDetector}, implements the network, and a further subclass
of it in turn (called \texttt{TelenetTrainer}) implements the special training procedure
required by VRD-RANS. Sampling probabilities or priors were not implemented
(unlike \cite{WANG202155}) in order to maximize the zero-shot perfomance of the model on fully unseen data --
these priors can also be seen as a way of artificially boosting performance by
knowing beforehand that the test split is sourced from the same dataset as the
training split, and thus both splits share similar statistical properties.
In our case this is counter-productive, as we are precisely trying to apply the
system to a transfer learning problem.

The hyperparameters used in this work are listed in Table~\ref{table:hyparam}.
The AdamW optimizer was selected instead of the classic Adam because it is capable
of performing weight normalization automatically, and thus it is possible to avoid
needing to explicitly specify normalization strategies in each layer. The rest
of the hyperparameters were found empirically.
The network was trained on a single NVIDIA Quadro RTX 5000 GPU.

The training process dynamically generates a minibatch for each image in the training set
(\texttt{prepare\_minibatch} method). The minibatch is generated by separately sampling
object pairs with ground truth predicate labels (\textit{positive}) and unlabelled object pairs
(\textit{negative}). The same number of positive and negative pairs are sampled. If this is not
possible due to there not being enough pairs of a certain kind, the minibatch is filled with
pairs of the other kind. If there are simply not enough pairs in total, it is filled with
random duplicate copies until the desired minibatch size is reached.

\subsection{Evaluation protocols for scene graph tasks}

Models dealing with scene graphs are known to be difficult to evaluate.
There exist several different tasks to evaluate them on, and it is
necessary to deal with problems arising from the incomplete/noisy/biased
nature of the datasets. In this section we detail the methodology we
devised to evaluate different approaches, as well as the challenges we faced.

First of all, existing work \cite{9157682,zellers2018scenegraphs}
considers and evaluates the following three
tasks separately, under various different names:

\begin{itemize}
	\item Predicate Detection (\textbf{PredDet}): Given an image and
	a list of objects in it (with bounding boxes and class information),
	rank all candidate relation triplets that can form a Scene Graph
	between the objects. This is the simplest version of the task
	and it is intended to only specifically evaluate the reliability
	of the relation detection.
	We chose to focus on evaluating this task.

	\item Visual Phrase Detection (\textbf{VPDet}): Given an image,
	detect all relation triplets that exist and assign them a single
	bounding box that covers the entire ``action.'' This was first
	popularised by \cite{lu2016visual}, however given the reliance on
	full object detectors by most models intended for generating scene
	graphs, we decided not to consider this task as it is a trivial
	variation of the more general scene graph generation task.

	\item Scene Graph Generation (\textbf{SGGen}): Given an image,
	detect all objects in it with bounding boxes and class information,
	and also all relation triplets between them in order to form a
	Scene Graph. This is the main task that a full model (incorporating
	an object detector) should aim to solve. Given the fact that the
	additional object detection phase introduces a new layer of noise
	and uncertainty, and in order to produce fair comparisons every model
	needs to be using the same object detector (which may or may not be
	possible), we also decided not to consider this task.
\end{itemize}

Much has been written about evaluation metrics for scene graphs.
The most widely used metric is Recall~@~K (\textbf{R@K}), which, as explained by
\cite{Tang_2019_CVPR,9157682},
has problems rooted in the heavily imbalanced distribution of relation
classes in datasets such as VG. For this reason, alternatives such as
the Zero-shot~Recall~@~K (\textbf{zR@K}) or Mean~Recall~@~K (\textbf{mR@K}) metric have
also been proposed.

In general, these metrics operate on an image by image basis, and they
involve ranking relation triplet predictions by their confidence score
generated by the network, and calculating the percentage of a set of
ground truth relation triplets that is covered by the top \textbf{K} selection of
predicted relation triplets. The metric for a given dataset is calculated
by averaging the metrics calculated on every suitable image in the dataset.
Recall was chosen as the base metric (as opposed to accuracy) because of the
incomplete/inexhaustive nature of the datasets used \cite{lu2016visual}.
Annotations in scene graph datasets do not exhaustively describe every single
object and every single relation between them. Using accuracy would result in
unfairly penalizing the network for possibly discovering new information about
the image that might have been missed by human labellers. For this reason,
the problem is approached as an information retrieval or ``search'' problem,
where the goal is returning relevant search results in response to a query.

The following is a summary of how each of these metrics work:

\begin{itemize}
	\item Recall~@~K (\textbf{R@K}): This is the base metric that
	calculates recall over all relation triplets in the ground truth.
	There is an extra implicit decision affecting the metric, which
	concerns how many highest scoring predicates to select for each object
	pair. Some authors consider picking the highest scoring predicate
	as the only one assigned to a given object pair
	\cite{lu2016visual,zellers2018scenegraphs}, other authors
	\cite{Tang_2019_CVPR,9157682} decided to select \textit{all} scores for
	all predicates, whereas some others \cite{vrdielkd2017,WANG202155}
	decided to make this an explicitly tunable \textit{graph constraint}
	hyperparameter \textbf{k} (lowercase, not to be confused with \textbf{K}).
	This hyperparameter is defined as the number of highest
	scoring predicates to select from each object pair.
	Given the multilabel nature of this problem, we decided to follow this
	last approach and explicitly report which different values are used
	for both \textbf{K} and \textbf{k}.

	\item Zero-shot~Recall~@~K (\textbf{zR@K}): This metric
	evaluates the network's ability to generalize its understanding of
	each predicate class by only evaluating the recall on the set of
	ground truth triplets involving object classes that have not appeared
	with corresponding predicates in the training set. As an example,
	\texttt{(person,laying~in,bed)} might appear in the training set, but
	\texttt{(cat,laying~in,bed)} might not. \textbf{zR@K} will ignore the former,
	but consider the latter as part of the ground truth set.

	\item Mean~Recall~@~K (\textbf{mR@K}): This metric is an attempt to
	solve the class imbalance problem by calculating \textbf{R@K} independently on
	each predicate class. In other words, the metric is subdivided
	into as many metrics as there are predicates. During the final
	aggregation step over the entire dataset, the individual \textbf{R@K} values
	of each image are aggregated separately for each predicate (note
	that the number of values in each group might be different, as some
	images might not contain examples of certain predicates, and
	thus they are not considered when calculating the \textbf{R@K} for said groups).
	The final \textbf{mR@K} value is the arithmetic mean of all \textbf{R@K} values
	calculated individually for each and every predicate.
\end{itemize}

In addition, we found that edge cases can arise during the calculation of
these metrics in certain situations, the handling of which we believe to
be important to fully disclose in order to enable fair comparisons between
results.

\begin{itemize}
	\item Sometimes, images have an empty ground truth triplet set. This can
	happen because the dataset simply does not record any relation triplets
	for a given image, or because there are no unseen triplet combinations
	during zero-shot metric calculation, or because a certain predicate
	does not appear in the image. We decided to simply skip the
	image during performance evaluation, since it is not possible to assign a metric
	to it, as the recall would involve a division by zero.

	\item The chosen \textbf{K} parameter could be lower than the number of triplets
	in the ground truth set of some image. In practice this should not happen with
	the sparse datasets we have, as they contain a low number of triplets
	per image. Nevertheless, we considered two possible solutions:

	\begin{itemize}
		\item Imposing a constraint on the value of \textbf{K} so that \textbf{K} is
		equal or greater than the size of the smallest non-zero ground
		truth set.
		\item Taking the minimum between \textbf{K} and the size of the ground
		truth set as the divisor when calculating recall. This is the
		solution we chose, because it does not make sense to calculate
		a metric in such way that full performance cannot be obtained.
	\end{itemize}

	\item Some object pairs in the ground truth set might have more predicates
	than the graph constraint hyperparameter \textbf{k}.
	We did not run into this situation because the test sets we used only have
	at most a single predicate in each labelled object pair.
	Nonetheless, we considered a solution, which is to calculate the
	size of the ground truth set in such way that no more than the given
	\textbf{k} are considered as the number of predicates accounted for evaluation
	within each labelled object pair.

	\item Some authors calculate the recall over the entire dataset instead
	of averaging the recall values of individual images. This has the effect
	of slightly underestimating the performance of the model, by giving
	greater weight to the contribution of images with a larger number of
	ground truth annotations. This was first notably done by \cite{lu2016visual},
	and followed by all papers that compare themselves against \cite{lu2016visual}.
	Subsequent works focused on other datasets \cite{Tang_2019_CVPR} opt for
	the more traditional way of aggregating values.
	We decided to follow existing practice to ensure fairness between results.

	\item Some authors only calculate recall over the set of object pairs
	that have corresponding label(s) in the ground truth, instead of taking
	the scores in all possible object pairs. This has the effect of inflating
	the reported recall values. This is most notably done when evaluating on
	the VRD dataset, for consistency with \cite{lu2016visual}.
\end{itemize}

\section{TERESA dataset experiments. Evaluating OG-SGG's transfer learning capabilities} \label{sec:teresa}

\begin{figure*}[t]
	\centering
	\insetfont

	\includegraphics[width=.65\linewidth]{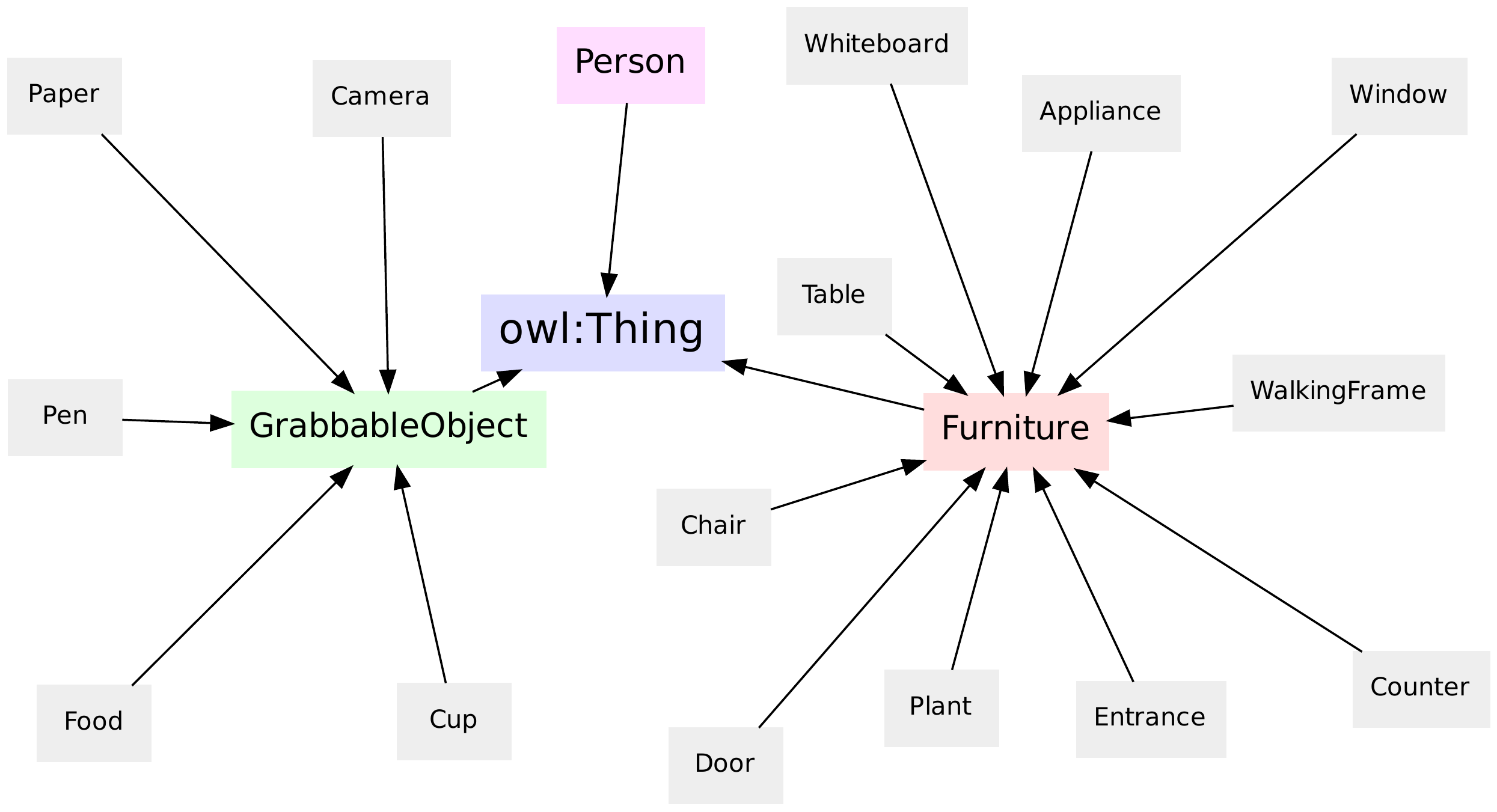}
	~
	\includegraphics[width=.3\linewidth]{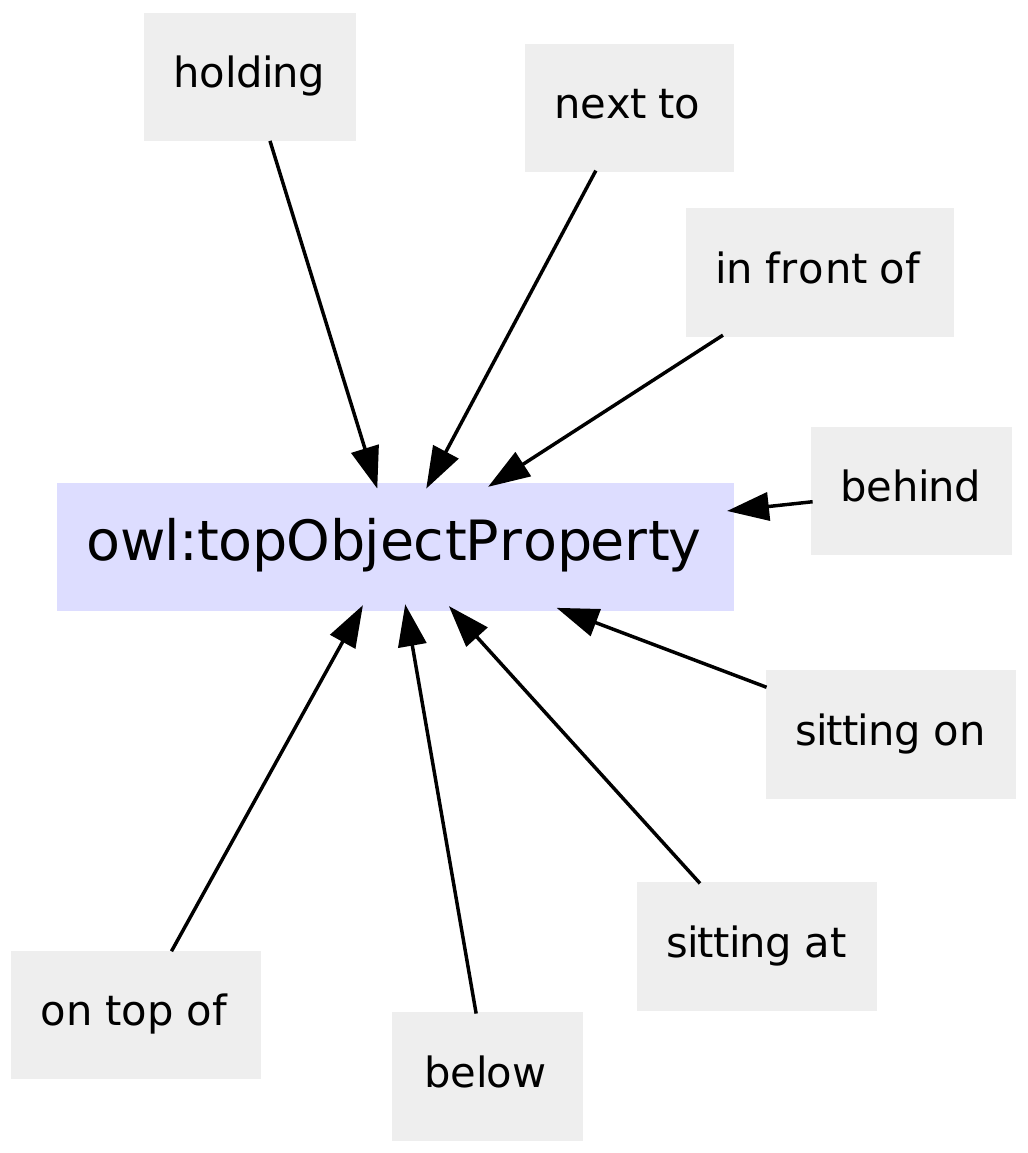}

	\caption{TERESA ontology, reflecting the dataset's main entities and
	relationships among them. The class hierarchy distinguishes \texttt{GrabbableObject}
	and \texttt{Furniture} in order to separate manipulable objects from
	pre-existing fixtures that form part of a room visited by the robot, whereas
	\texttt{Person}, the main focus of a telepresence robotics application, is the domain of
	certain special purpose predicates such as \texttt{holding} or
	\texttt{sitting at/on}.}
	\label{fig:onto}
\end{figure*}

\begin{table*}[tb!]
	\centering
	\insetfont
	\caption{TERESA dataset statistics. Training datasets are reported in their
	original base form, their (e.g., domain-) filtered, and their filtered + (ontological axioms-) augmented form.}
	\begin{tabular}{ |r|c|c|c||c|c|c||c| }
		\cline{2-8}
		\multicolumn{1}{c|}{} & \multicolumn{3}{c||}{VG-SGG (Train)} & \multicolumn{3}{c||}{VG-indoor (Train)} & \multicolumn{1}{c|}{TERESA} \\
		\cline{2-7}
		\multicolumn{1}{c|}{} & Base & Filter & F.+Aug. & Base & Filter & F.+Aug. & (Test) \\
		\hline
		Number of images          & 62723 & 51162 & 51162 & 3036 &  2505 &  2505 &     25 \\
		Connected objects/image   &  6.73 &  4.59 &  4.60 & 6.16 &  4.31 &  4.31 &  10.42 \\
		Triplets/image            &  5.46 &  5.91 &  6.02 & 4.72 &  5.27 &  5.33 &  21.46 \\
		Annotated pairs/image     &  5.15 &  5.79 &  5.89 & 4.48 &  5.19 &  5.25 &  21.38 \\
		\% pairs with annotations &  9.11 & 22.68 & 23.04 & 9.18 & 24.61 & 24.90 &  18.05 \\
		\hline
	\end{tabular}
	\label{table:dataset_stats}
\end{table*}

\begin{table*}[tb!]
	\centering
	\insetfont
	\caption{Evaluation results on TERESA test set. The model was trained on
	6 different dataset splits (3 for each source training dataset), and evaluated
	with and without post-processing (\textit{``Post''} column). The different splits are intended to
	test the efficacy of the filtering and data augmentation techniques proposed
	in this work. The different metrics are computed for different top \textbf{K}
	predicted relation triplets (20, 50 and 100) per image and different graph
	constraint hyperparameter \textbf{k} values (1 and 8 predicates per object pair).
	The best results for each source training dataset are marked in bold.}
	\begin{tabular}{ |l|c|c c c|c c c||c c c|c c c| }
		\cline{3-14}
		\multicolumn{2}{c|}{} & \multicolumn{12}{c|}{Metrics for Predicate Detection (PredDet)} \\
		\hline
		\multicolumn{1}{|c|}{\multirow{2}{*}{Dataset}} & \multirow{2}{*}{Post}
			& \multicolumn{3}{c|}{\textbf{R@K} (\textbf{k} = 1)} & \multicolumn{3}{c||}{\textbf{R@K} (\textbf{k} = 8)} & \multicolumn{3}{c|}{\textbf{mR@K} (\textbf{k} = 1)} & \multicolumn{3}{c|}{\textbf{mR@K} (\textbf{k} = 8)} \\
			& & 20 & 50 & 100 & 20 & 50 & 100 & 20 & 50 & 100 & 20 & 50 & 100 \\
		\hline
		\multirow{2}{*}{VG-SGG unmodified dataset}
			& \N & 27.0 & 34.7 & 41.9 & 23.8 & 34.8 & 51.1 & 19.1 & 30.6 & 36.4 & 29.3 & 42.7 & 57.0 \\
			& \Y & 28.4 & 36.0 & 43.2 & 29.5 & 42.2 & 57.9 & 32.4 & 44.6 & 51.1 & 33.5 & 49.7 & \textbf{63.0} \\
		\hline
		\multirow{2}{*}{VG-SGG with filtering}
			& \N & 40.0 & 43.4 & 47.7 & 42.0 & 49.0 & 60.0 & 42.2 & 48.8 & 50.9 & 43.5 & 51.5 & 57.4 \\
			& \Y & 44.7 & 47.9 & 53.2 & \textbf{46.5} & 53.4 & 66.3 & 44.0 & 51.2 & \textbf{53.6} & 44.7 & \textbf{53.9} & 60.5 \\
		\hline
		\multirow{2}{*}{VG-SGG with filtering/augmentation}
			& \N & 39.8 & 43.9 & 48.6 & 40.7 & 49.8 & 61.5 & 42.5 & 47.1 & 50.6 & 43.4 & 51.2 & 57.9 \\
			& \Y & \textbf{44.9} & \textbf{49.3} & \textbf{54.0} & \textbf{46.5} & \textbf{54.0} & \textbf{68.3} & \textbf{44.6} & \textbf{49.9} & 53.3 & \textbf{45.2} & 53.0 & 61.7 \\
		\hline
		\hline
		\multirow{2}{*}{VG-indoor unmodified dataset}
			& \N & 26.1 & 33.7 & 40.2 & 26.0 & 41.4 & 56.1 & 10.5 & 20.6 & 25.2 & 19.8 & 35.5 & 53.6 \\
			& \Y & 30.7 & 38.4 & 45.2 & 32.7 & 45.6 & 59.6 & 22.0 & 39.5 & 44.7 & 23.3 & 39.5 & 58.1 \\
		\hline
		\multirow{2}{*}{VG-indoor with filtering}
			& \N & 41.2 & 41.3 & 46.7 & 42.2 & 48.1 & 59.2 & 43.7 & 50.6 & 45.5 & 44.7 & 56.0 & 65.8 \\
			& \Y & 43.6 & 45.3 & 51.8 & \textbf{44.2} & \textbf{51.5} & \textbf{62.8} & 45.0 & 52.9 & 59.6 & \textbf{47.1} & 57.7 & \textbf{69.3} \\
		\hline
		\multirow{2}{*}{VG-indoor with filtering/augmentation}
			& \N & 42.1 & 42.6 & 46.9 & 42.2 & 48.5 & 58.1 & 45.5 & 53.0 & 56.1 & 46.2 & 56.8 & 63.8 \\
			& \Y & \textbf{44.4} & \textbf{46.1} & \textbf{51.9} & 43.3 & \textbf{51.5} & 62.0 & \textbf{46.8} & \textbf{54.8} & \textbf{61.0} & 46.6 & \textbf{58.3} & 66.6 \\
		\hline
	\end{tabular}
	\label{table:teresa}
\end{table*}

\begin{figure*}[tb!]
	\centering
	\insetfont
	\begin{tabular}{ c c }
		\includegraphics[width=.4\linewidth]{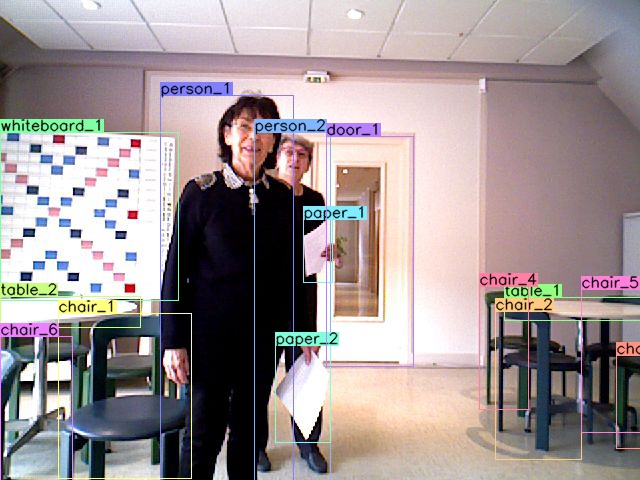} &
		\includegraphics[width=.4\linewidth]{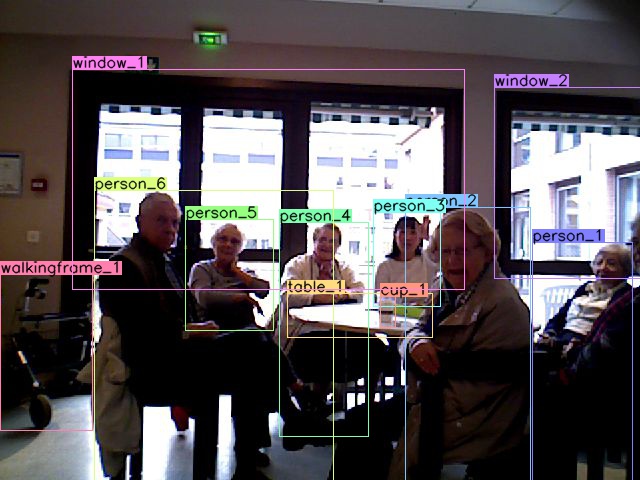} \\
		\includegraphics[width=.4\linewidth]{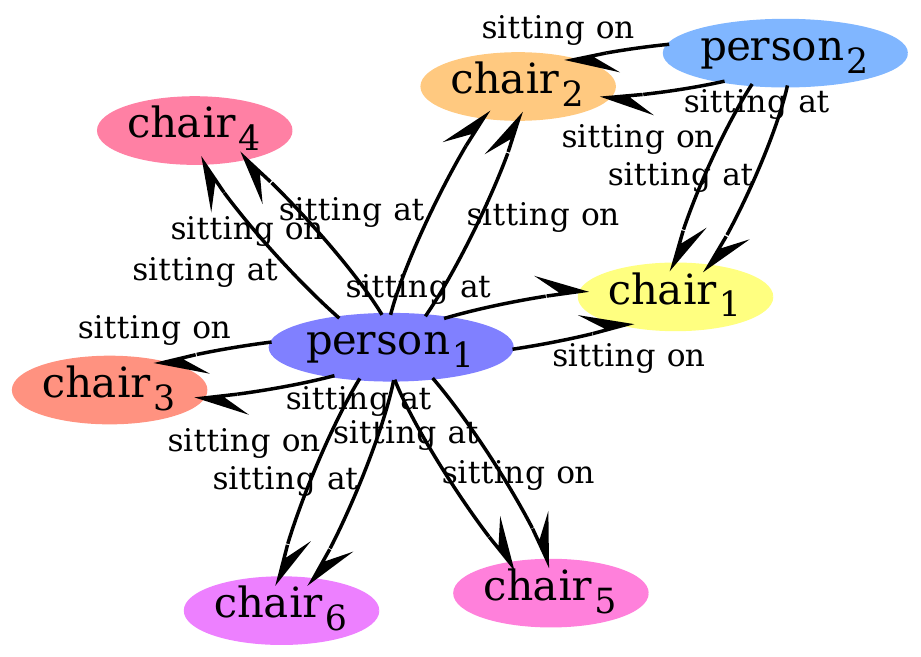} &
		\includegraphics[width=.25\linewidth]{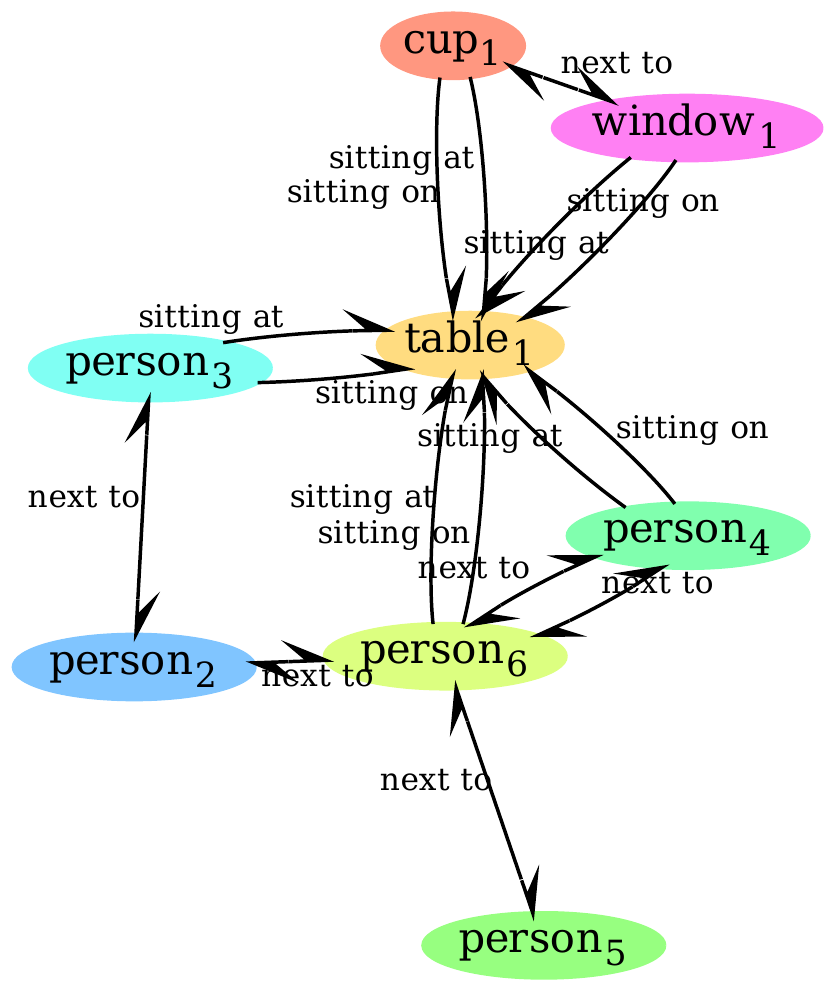} \\
		\pbox{.4\linewidth}{\centering
			\textit{
				Person 1 is sitting on chairs 1, 2, 3, 4, 5 and 6.\newline
				Person 2 is sitting on chairs 1 and 2.
			}
		} &
		\pbox{.4\linewidth}{\centering
			\textit{
				Persons 3, 4, 5 are sitting at table 1.\newline
				Person 2 is next to Person 3 and 4.\newline
				Person 3 is next to Person 2.\newline
				Person 4 is next to Person 6.\newline
				Person 5 is next to Person 6.\newline
				Person 6 is next to Person 2 and 4.
			}
		} \\
		\includegraphics[width=.4\linewidth]{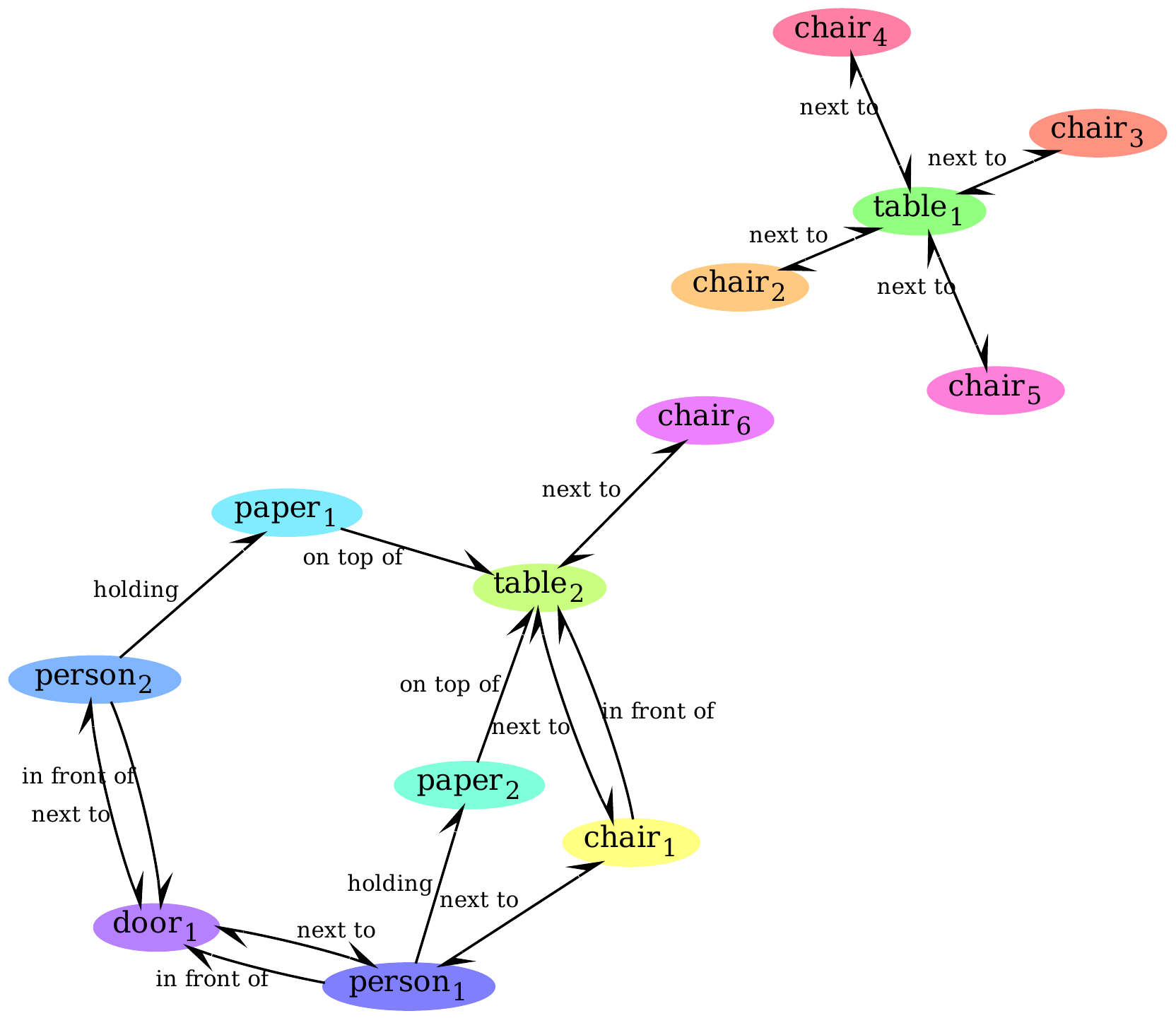} &
		\includegraphics[width=.3\linewidth]{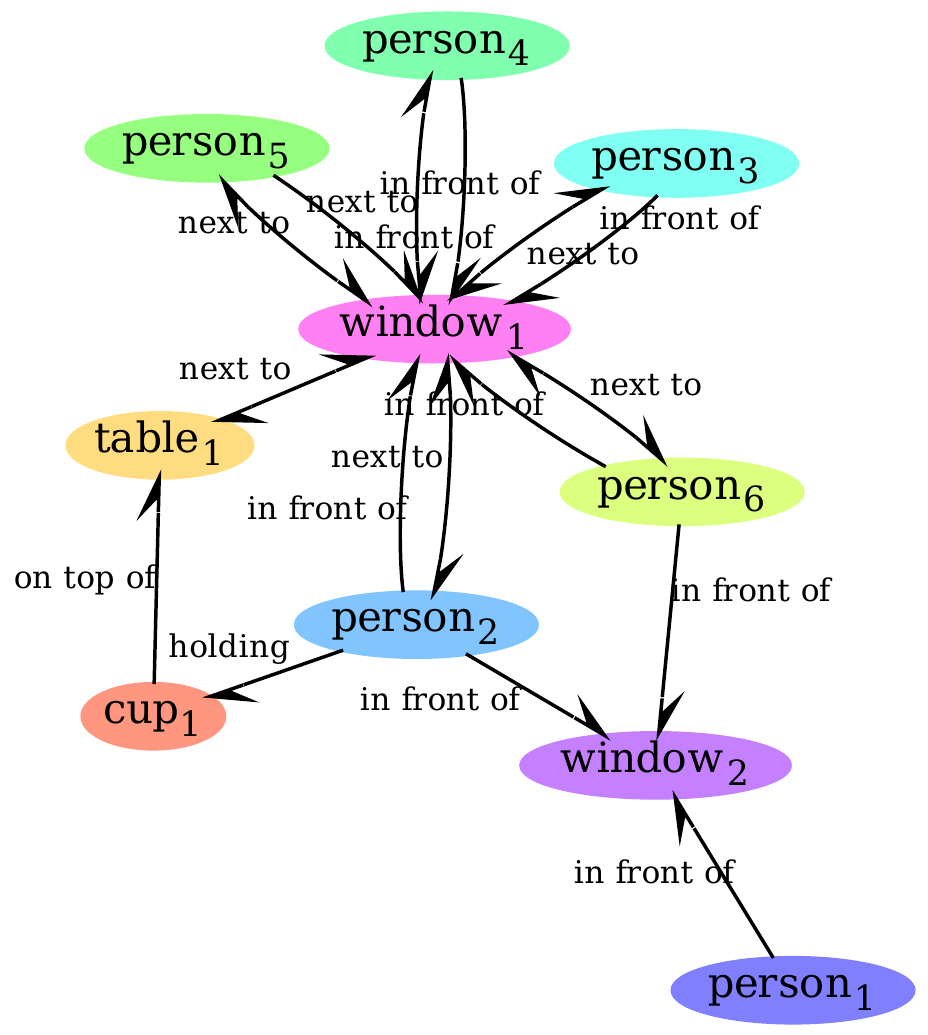} \\
		\pbox{.4\linewidth}{\centering
			\textit{
				Person 1 and 2 are in front of a door.\newline
				Person 1 is holding a piece of paper.\newline
				Person 1 is next to chair 1.\newline
				Person 2 is holding a piece of paper.
			}
		} &
		\pbox{.4\linewidth}{\centering
			\textit{
				Persons 1 and 2 are in front of a window.\newline
				Persons 2, 3, 4, 5 and 6 are in front of a window.\newline
				Person 2 is holding a cup.
			}
		} \\
	\end{tabular}
	\caption{
		Qualitative results on TERESA test set, gathered using the TERESA robot of Fig. \ref{fig:teresa}.
		\textbf{Top row}: Image with object annotations.
		\textbf{Middle row}: ``VG-SGG baseline without OG-SGG.''
		\textbf{Bottom row}: ``VG-SGG with full stack OG-SGG.''
	}
	\label{fig:qualit}
\end{figure*}

In order to evaluate the effects of the ontology-guided scene graph generation
(OG-SGG) framework, we applied it to a telepresence robotics use case.
Specifically, we utilized data from the TERESA \cite{teresa_icra2015} European Project,
which involved a telepresence robot being used within an elderly day-care centre.
The robot is used by both residents and caregivers in order to remotely
connect and interact with other people in the centre's cafeteria
(see Figs. \ref{fig:teresa} or \ref{fig:examplegraph} for some examples).
13 sessions were carried out in total, during which large amounts of data were collected from the robot's cameras and other sensors. For the purposes of these experiments, a small sample of 25 images were extracted and manually annotated with the objects and relationships present within them.

The end goal of this experiment is determining whether these techniques allow us
to transfer learned knowledge from existing datasets into a completely new
problem domain, with minimal work put into defining the rules required to
perform this conversion. We also aim to achieve the double-sided benefit of
refining the VQA tasks that allow robotic agents to automatically reason
about user input, e.g. users of the telepresence system may want to ask about
the locations of objects or people, or refer to them by their relation to
other entities in the scene.
In order to enable the reproducibility of this work, we published the
entirety of the source code we developed, along with the TERESA dataset
\footnote{\url{https://github.com/robotics-upo/og-sgg}}.

We created a simple ontology using Protégé \cite{protege2018},
and annotated all objects on the images
with bounding boxes, as well as the corresponding relation triplets.
This ontology, although fairly simple, nevertheless encompasses
all the important objects and relations of the application scenario.
The scheme of the ontology can be seen in Figure~\ref{fig:onto}.
The ontology was validated using the FaCT++ \cite{factplusplus} reasoner.
Several top level \textit{classes} were defined: \texttt{Furniture} (stationary
objects present in a room), \texttt{GrabbableObject} (objects that can be grabbed
and moved by the robot) and \texttt{Person} (corresponding to humans).
A few \textit{object properties} were also defined, corresponding to common
relationships between entities in a scene graph. For example, \texttt{on top of}
is a Functional object property with the domain \texttt{Appliance or GrabbableObject}
and the range \texttt{Appliance or Counter or Table}. This defines several axioms:
1) something can only be on top of a single surface (not several at the same time),
2) only appliances (e.g. microwaves) or small, movable objects can be on top, whereas
3) said acceptable ``surfaces'' can only be tables, counters or other appliances.
Another example is \texttt{sitting on}: a Functional and InverseFunctional property
with the domain \texttt{Person} and the range \texttt{Chair}. This enforces that
people can only be sitting on a single chair, and a chair cannot host multiple people
at the same time.

We also performed an ablation study of these techniques, for the purpose of which
we prepared six different training splits for the network. In each split,
we tried combinations of input datasets as well as enabling or disabling parts
of the dataset filtering/augmentation logic (note that the augmentation logic
needs predicates to be already adapted to the ontology, that is, the filtering
logic needs to be done previously). We tested VG-SGG's training split,
as well as a filtered version of it only containing images classified as
``indoors'' by GQA, which we called VG-indoor. This filtered subset, after
being processed with the ontology guided procedure, contains around 2500
training images (out of which around 250 are reserved for validation and
hyperparameter tuning), and is intended to test OG-SGG in cases where the
source dataset is considerably smaller in size and scope.

Table~\ref{table:dataset_stats} shows several statistics about the different
dataset splits used for training, as well as our custom TERESA test set used
for evaluation. We report the average number of objects per image that are
connected by relation triplets, which naturally decreases the more filtering
is done. On the other hand, the average number of relation triplets increases
even prior to applying the ontology guided data augmentation process. We hypothesize
this to be caused by the removal of especially noisy or underlabelled images,
itself a side effect of the ontology guided filter.
The data augmentation process causes a milder increase than expected, probably due
to the relative simplicity of the ontological model we designed for TERESA.
Similar things can also be said about the
other metrics, such as the average number of object pairs that are annotated
with predicates, or the average percentage of such pairs in the image.
This last metric is intended to measure the degree of annotation density (or
rather, sparsity) in the triplet annotations by taking the set of objects
connected by relation triplets, and expressing the number of pairs with
annotations as a percentage of the total number of possible pairs within the set.

\subsection{Quantitative results}

Table~\ref{table:teresa} reports evaluation results on the TERESA test set
for the network trained on each dataset split, as well as with and without the
ontology founded output post-processing procedure.
We also emphasize that no images from TERESA were used during training.
The metrics for the splits corresponding to unmodified source training datasets were computed
by first adapting the output of the model to fit the predicates in the ontology
with minimal post-processing and no ontological filtering.
Specifically,
each predicate in the ontology was assigned an output score of the average of
the predicate scores corresponding to its mapped set of predicates (the same
used during dataset filtering).
Although we report \textbf{R@K} metrics, the network is trained on a
completely different dataset to the one used for testing; and the set of semantic vectors
corresponding to object classes seen by the network during training is also entirely
different from the one provided during testing -- in other words, the metrics are
calculated purely on zero-shot triplets never seen during training,
thus effectively calculating a sort of \textbf{zR@K}.
It can be readily observed that using a training dataset
specifically prepared to target a desired set of predicates uplifts performance.

\subsubsection{Choice of training dataset}

The choice of original dataset onto which OG-SGG is applied also produced interesting results.
The results for VG-indoor trained models are highly competitive against those
trained on regular VG-SGG.
In some metrics such as \textbf{mR@K} it outperforms VG-SGG trained models, with
the full stack OG-SGG version of its model taking the performance crown overall.
Even more, the model trained on VG-indoor with full stack OG-SGG outperforms
the VG-SGG baseline without OG-SGG, despite having seen over twenty times less images
during the training process.

On the other hand, models trained on VG-SGG with full stack OG-SGG dominate in \textbf{R@K}.
This might be caused by the
previously explained problem caused by predicate bias in \textbf{R@K}. Specifically,
\textbf{mR@K} attempts to paint a more balanced picture by weighing the importance of
all predicates equally in its formula, therefore allowing the tails of the
predicate frequence distribution to have a fair say in the result.
Thus, it could be said that training on VG-indoor helped the model generalize better
on the less frequent predicates.

\subsubsection{Ablation study}

\textbf{Post-processing}. The ablation study reveals the great importance of the post-processing stage,
which enforces the axioms defined in the ontology, purging lower-ranking
inconsistent triplet proposals,
and thus results in increased performance across the board. An interesting
observation can be made for the improvements obtained when training on unmodified datasets
(i.e. when no other components of OG-SGG are used), which bring the metrics close
to those observed in filtered datasets (with no post-processing stage).
We, as a result, believe this component to be the most significant contribution of this work.

\textbf{Filtering}. As previously mentioned, filtering the dataset with the ontology is also majorly
responsible for the improved performance. In other words, this process optimizes
the transfer learning capabilities of scene graph generation networks, allowing
users to obtain better results by ``recycling'' existing datasets. Specifically,
it can be seen that filtering brings significant boosts to \textbf{mR@K}, which is
indicative of greater generalization capability. \textbf{R@K} also receives a boost,
although it is not as dramatic in comparison.

\textbf{Augmentation}. On the other hand, the extra training data produced
by the augmentation does not seem to have produced a significant improvement
as hoped, i.e. it could be said to lie within the margin of error caused by
the variability of the training or stratification processes.
This could be caused by not enough complexity/richness in the definition of the
TERESA ontology.
Nonetheless, we still consider it relevant to continue researching more
robust ways of augmenting existing datasets using ontological reasoning.

\subsection{Qualitative results}

Fig.~\ref{fig:qualit} shows two selected qualitative examples. The graphs
were generated by running the images through the model and picking the 16
highest scoring generated triplets. In the case of the version with
post-processing, adding a triplet also adds all associated implicit triplets.
Additionally, disallowed triplets, as well as triplets that were previously
added implicitly, are not considered in the score ranking, meaning they do not
count towards the triplet limit. In the baseline generated graphs, some
violations can be spotted (such as multiple people sitting on multiple chairs,
or windows sitting at tables).
On the other hand, the graphs generated with the proposed techniques in this work have some
discernible structures, such as people holding objects, or chairs being
next to tables.

With this said, some
shortcomings can be seen, such as the generator being unable to tell if an
object is being held by a person or is located ``on top'' of a certain table.
In these situations, the generator simply asserts both of these possibilities.
In addition, some potentially useful information such as proximity relations
between people seems to be deemphasized in favor of other structures
that the network was able to learn with the same predicates.
Likewise, the network fails to learn cues for discerning the various levels
of depth present in images, resulting in the understanding of proximity
relationships being reduced to mere 2D spatial proximity, as can be seen in
how objects close to people's hands are nearly always detected as being held.
This indicates that more input (such as Depth information) might be necessary
for the network, and that higher order ontological rules need to be implemented
in order to decide between different possibilities. Nonetheless, there is still
a noticeable improvement compared to baseline non-ontology-guided methods.

Sample textual representations of each scene graph revolving around detected
people were also generated by enumerating all triplets that have \texttt{Person}
entities as their subject. These representations are intended to be an example
of downstream automatic scene captioning tasks that are common in telepresence
robotics.

\section{Additional experiments using AI2THOR} \label{sec:ai2thor}

\begin{table*}[t]
	\centering
	\insetfont
	\caption{Dataset statistics for AI2THOR. Training datasets derived from VG are reported in their
	original base form, their (e.g., domain-) filtered, and their filtered + (ontological axioms-) augmented form.}
	\begin{tabular}{ |r|c|c|c||c|c|c||c| }
		\cline{2-8}
		\multicolumn{1}{c|}{} & \multicolumn{3}{c||}{VG-SGG (Train)} & \multicolumn{3}{c||}{VG-indoor (Train)} & AI2THOR \\
		\cline{2-7}
		\multicolumn{1}{c|}{} & Base & Filter & F.+Aug. & Base & Filter & F.+Aug. & (Test) \\
		\hline
		Number of images          & 62723 & 34585 & 34585 & 3036 &  1679 &  1679 &   113 \\
		Connected objects/image   &  6.73 &  3.83 &  3.84 & 6.16 &  3.67 &  3.67 & 12.42 \\
		Triplets/image            &  5.46 &  4.82 &  4.84 & 4.72 &  4.36 &  4.38 & 48.43 \\
		Annotated pairs/image     &  5.15 &  4.78 &  4.80 & 4.48 &  4.31 &  4.33 & 45.11 \\
		\% pairs with annotations &  9.11 & 28.40 & 28.51 & 9.18 & 26.83 & 26.90 & 27.27 \\
		\hline
	\end{tabular}
	\label{table:ai2thor_dataset_stats}
\end{table*}

\begin{table*}[tb!]
	\centering
	\insetfont
	\caption{Evaluation results on AI2THOR test set. The model was trained on
	6 different dataset splits (3 for each source training dataset), and evaluated
	with and without post-processing (\textit{``Post''} column).}
	\begin{tabular}{ |l|c|c c c|c c c||c c c|c c c| }
		\cline{3-14}
		\multicolumn{2}{c|}{} & \multicolumn{12}{c|}{Metrics for Predicate Detection (PredDet)} \\
		\hline
		\multicolumn{1}{|c|}{\multirow{2}{*}{Dataset}} & \multirow{2}{*}{Post}
			& \multicolumn{3}{c|}{\textbf{R@K} (\textbf{k} = 1)} & \multicolumn{3}{c||}{\textbf{R@K} (\textbf{k} = 8)} & \multicolumn{3}{c|}{\textbf{mR@K} (\textbf{k} = 1)} & \multicolumn{3}{c|}{\textbf{mR@K} (\textbf{k} = 8)} \\
			& & 20 & 50 & 100 & 20 & 50 & 100 & 20 & 50 & 100 & 20 & 50 & 100 \\
		\hline
		\multirow{2}{*}{VG-SGG unmodified dataset}
			& \N & 19.1 & 24.4 & 32.1 & 21.9 & 27.2 & 40.5 &  7.0 & 11.8 & 16.2 &  9.4 & 15.2 & 24.6 \\
			& \Y & \textbf{21.0} & \textbf{29.7} & \textbf{40.4} & \textbf{23.7} & 32.7 & 48.3 &  7.5 & 13.5 & 18.3 & 10.1 & 17.6 & 28.2 \\
		\hline
		\multirow{2}{*}{VG-SGG with filtering}
			& \N & 16.7 & 27.5 & 37.1 & 17.2 & 28.0 & 44.3 & 11.5 & 20.6 & 27.3 & 13.0 & 24.4 & 42.4 \\
			& \Y & 17.3 & 28.7 & 38.7 & 18.9 & \textbf{33.1} & \textbf{52.5} & 11.6 & 21.1 & \textbf{27.9} & \textbf{13.8} & 28.0 & \textbf{48.3} \\
		\hline
		\multirow{2}{*}{VG-SGG with filtering/augmentation}
			& \N & 19.3 & 27.7 & 36.8 & 19.7 & 28.3 & 43.4 & 12.3 & 20.5 & 26.6 & 13.2 & 26.0 & 40.4 \\
			& \Y & 20.0 & 29.0 & 38.5 & 20.9 & 32.4 & 50.7 & \textbf{12.9} & \textbf{21.2} & 27.2 & \textbf{13.8} & \textbf{29.0} & 45.3 \\
		\hline
		\hline
		\multirow{2}{*}{VG-indoor unmodified dataset}
			& \N & 18.7 & 21.4 & 26.0 & 22.1 & 26.5 & 37.4 &  9.3 & 17.7 & 23.0 & 12.2 & 22.3 & 31.5 \\
			& \Y & \textbf{22.6} & \textbf{27.8} & \textbf{34.5} & \textbf{25.3} & 31.1 & 45.1 & 10.8 & 20.7 & 25.7 & 13.5 & 24.8 & 35.8 \\
		\hline
		\multirow{2}{*}{VG-indoor with filtering}
			& \N & 20.5 & 23.4 & 28.1 & 23.2 & 28.4 & 39.0 & 14.0 & 22.7 & 28.4 & 16.0 & 28.8 & 42.6 \\
			& \Y & 21.4 & 25.7 & 31.4 & 24.7 & \textbf{32.9} & \textbf{45.9} & \textbf{14.5} & \textbf{24.1} & \textbf{30.8} & \textbf{16.9} & \textbf{32.5} & \textbf{48.0} \\
		\hline
		\multirow{2}{*}{VG-indoor with filtering/augmentation}
			& \N & 19.3 & 22.2 & 26.2 & 21.1 & 27.7 & 38.2 & 13.6 & 21.6 & 26.2 & 15.1 & 28.2 & 40.7 \\
			& \Y & 20.3 & 23.9 & 29.5 & 22.5 & 30.9 & 43.2 & 13.9 & 22.3 & 28.0 & 15.8 & 30.8 & 43.9 \\
		\hline
	\end{tabular}
	\label{table:ai2thor}
\end{table*}

\begin{figure*}[tb!]
	\centering
	\insetfont
	\begin{tabular}{ c c }
		\includegraphics[width=.4\linewidth]{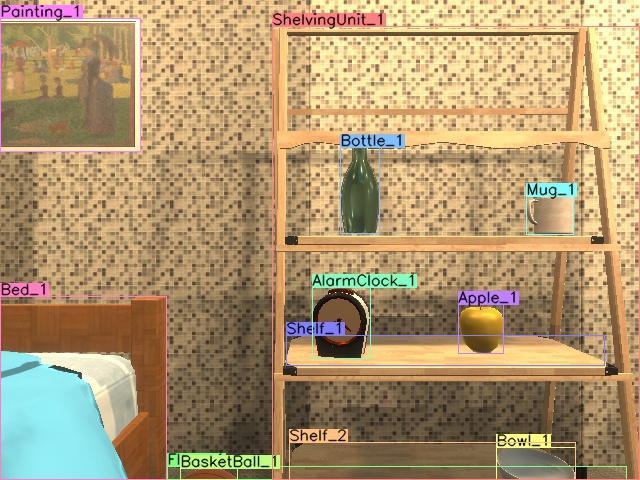} &
		\includegraphics[width=.4\linewidth]{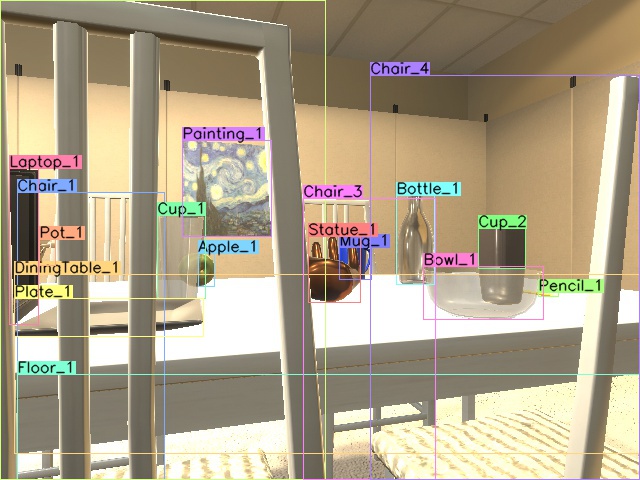} \\
		\includegraphics[width=.35\linewidth]{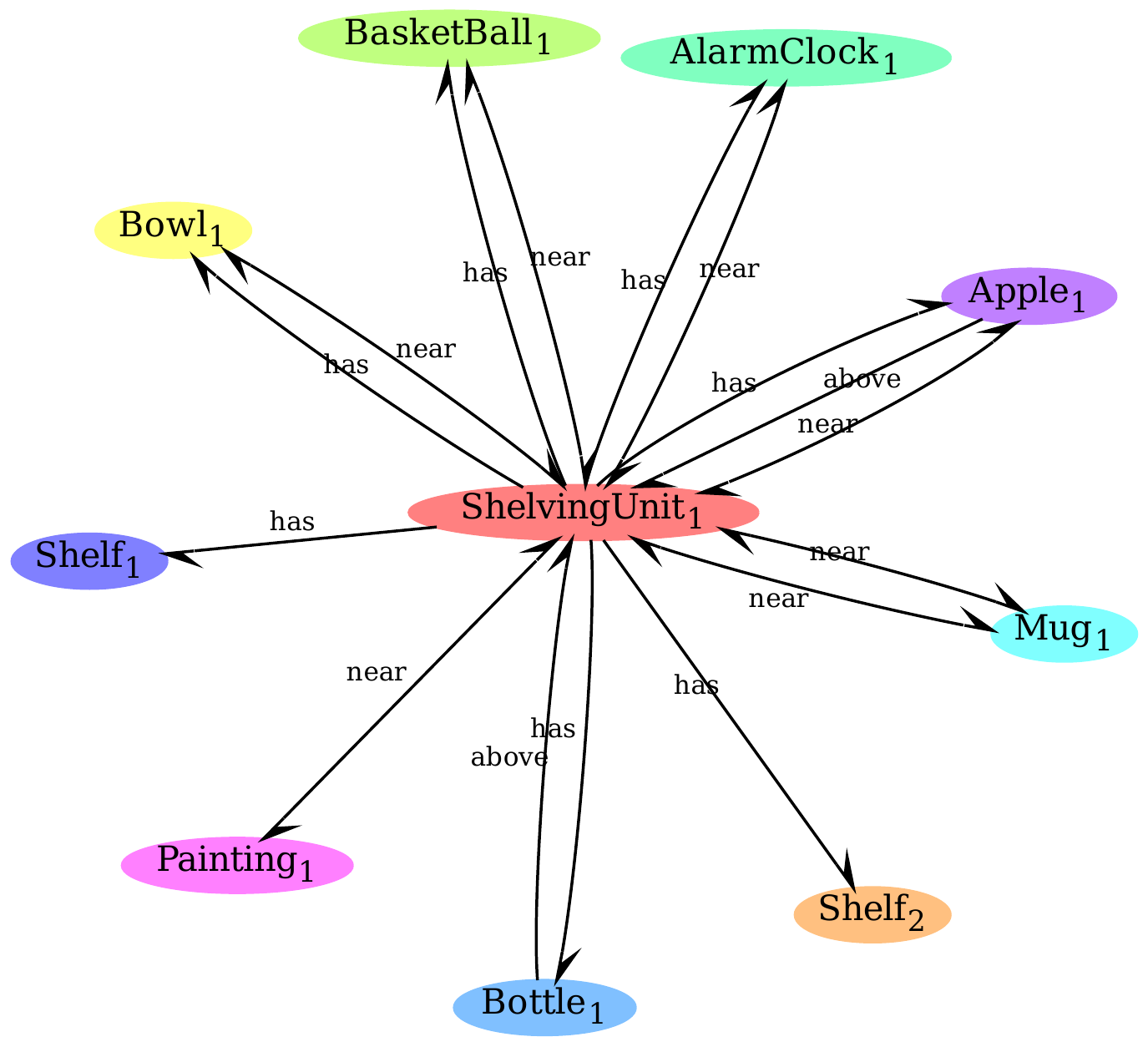} &
		\includegraphics[width=.4\linewidth]{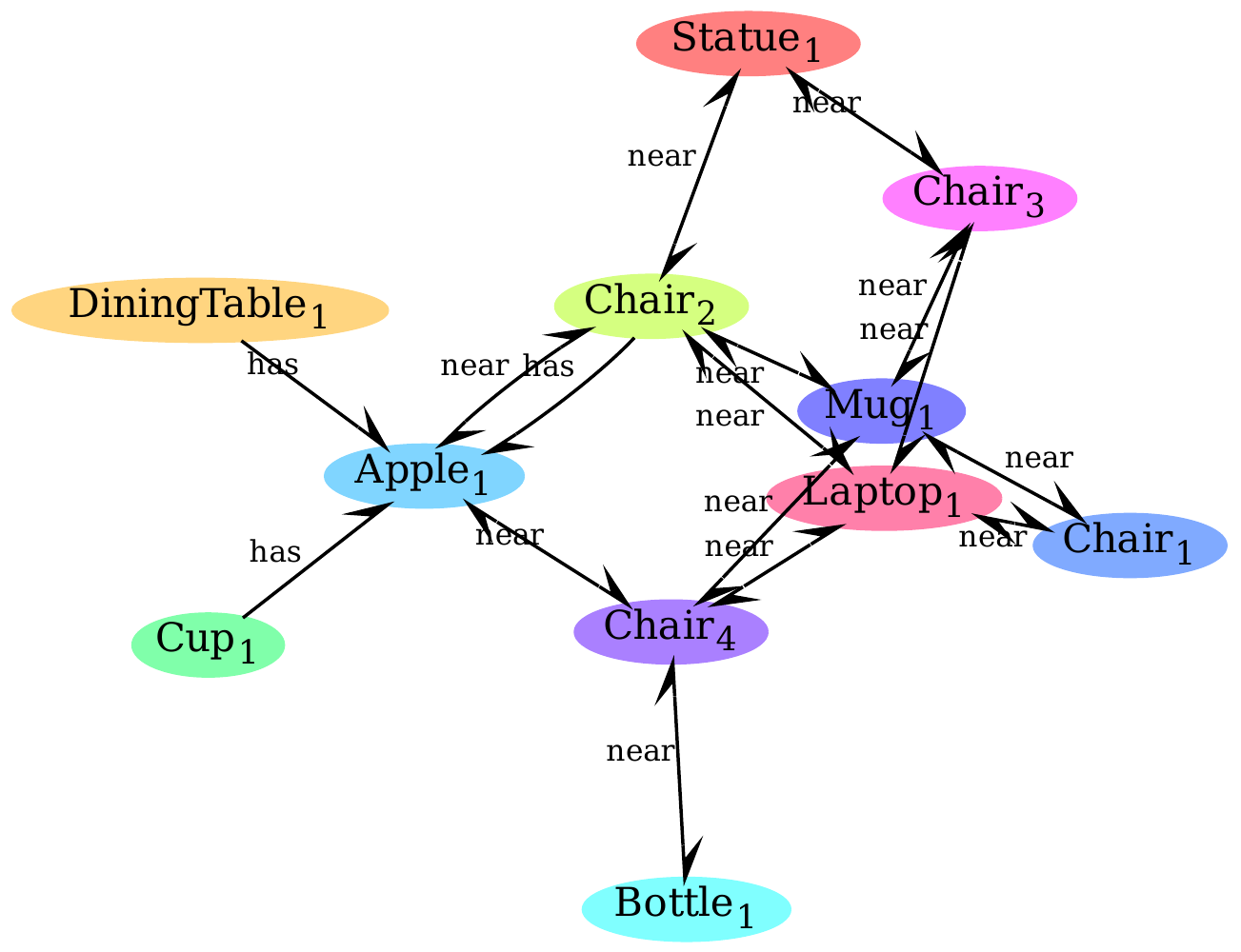} \\
		\includegraphics[width=.3\linewidth]{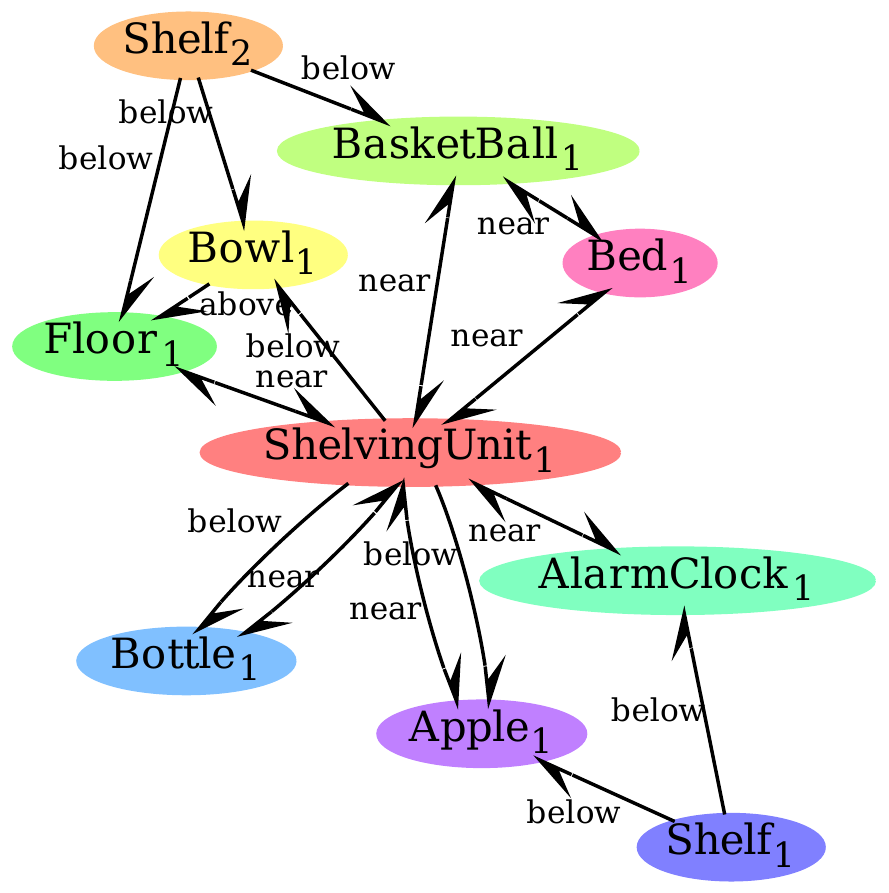} &
		\includegraphics[width=.4\linewidth]{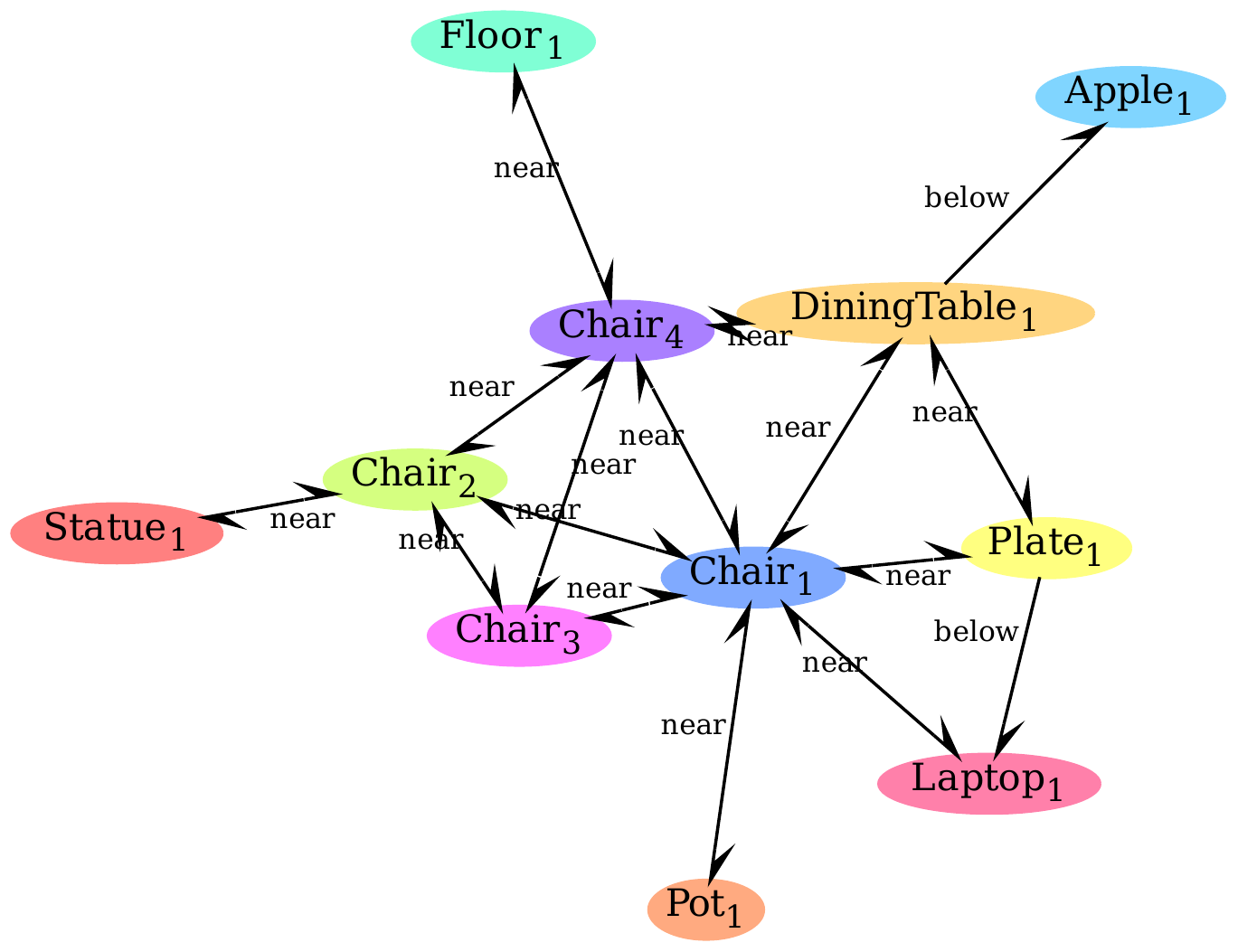} \\
	\end{tabular}
	\caption{
		Qualitative results on AI2THOR test set.
		\textbf{Top row}: Image with object annotations.
		\textbf{Middle row}: ``VG-SGG baseline.''
		\textbf{Bottom row}: ``VG-SGG with filter + post processing.''
	}
	\label{fig:ai2thor_qualit}
\end{figure*}

We decided to apply OG-SGG to a similar but different robotics scenario.
Specifically, the AI2THOR framework \cite{ai2thor} is a near photo-realistic
interactable framework for embodied AI agents, with the goal of facilitating
the creation of visually intelligent models and pushing the research forward
in that domain. One of the environments present in this framework is
RoboTHOR \cite{robothor} which has a specific focus on simulation-to-real transfer
of visual AI (especially semantic navigation).

The AI2THOR system provides a simulation framework, from which different
synthesized inputs from the robot can be extracted and actions can be
subsequently fed to the robot. In order to test OG-SGG, we collected 113 images in total
from the 3 RoboTHOR validation scenes, in all 5 different configurations.
Each image was captured using different material randomization, and the
camera was moved so that at least 6 objects are visible within the view.
Afterwards, ground truth scene graphs were automatically generated with
fixed, hard-coded rules that leverage information about the objects
provided directly by the system, such as absolute positions or container
relationships. Another simple ontology was also created to
ground the concepts present in these scene graphs.

We performed the same experiments carried out for the TERESA dataset, including
the ablation study. Table~\ref{table:ai2thor_dataset_stats} contains dataset
statistics, which show a similar increase in dataset density after applying
filtering, which brings it to a level close to that of the desired target dataset.
On the other hand, the data augmentation process seems to not result in a significant
increase in density. This is probably once again due to lack of higher level rules
in the ontology that can deduce new triplets from existing ones.
Table~\ref{table:ai2thor} likewise confirms the performance improvements brought
by the filtered dataset, and the post-processing stage. The metrics corresponding
to the model trained on augmented splits seem to be within margin of error compared
to ontology-guided filtering without data augmentation -- this is consistent with the
aforementioned lack of increase in density.
Interestingly, the choice of filtered indoor dataset here has clearly resulted in
worse performance across the board, except in the baseline comparison case.
This might be due to the harsher filtering caused by the AI2THOR ontology, which
has effectively halved the number of images available and thus might not be able
to form a critical mass necessary for the network to be successfully trained.

Fig.~\ref{fig:ai2thor_qualit} showcases a few qualitative examples, taking the
VG-SGG split with filtering and no data augmentation as the basis for the trained model.
It can be seen that in general the network has a tendency to be overly enthusiastic
in predicting \texttt{has} and \texttt{near}, the former in the baseline and the
latter in our improved system. In any case, it can be observed that ontology violations
(such as (\texttt{Cup}\_1,\texttt{has},\texttt{Apple}\_1) in the second image) do not
appear in our improved output thanks to the post-processing.

\section{Additional experiments using different models} \label{sec:othermodel}

\begin{figure*}[t!]
	\centering
	\insetfont

	\includegraphics[width=.4\linewidth]{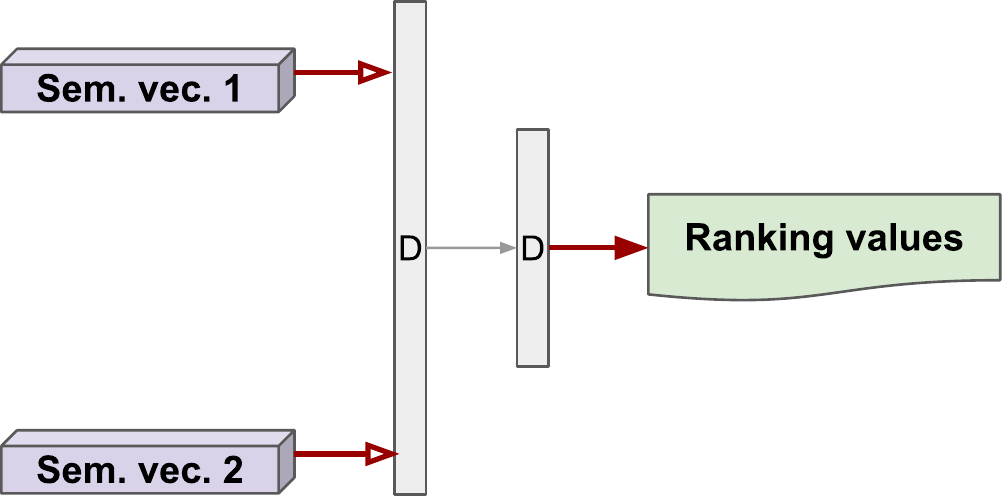}
	\hspace{7mm}
	\includegraphics[width=.4\linewidth]{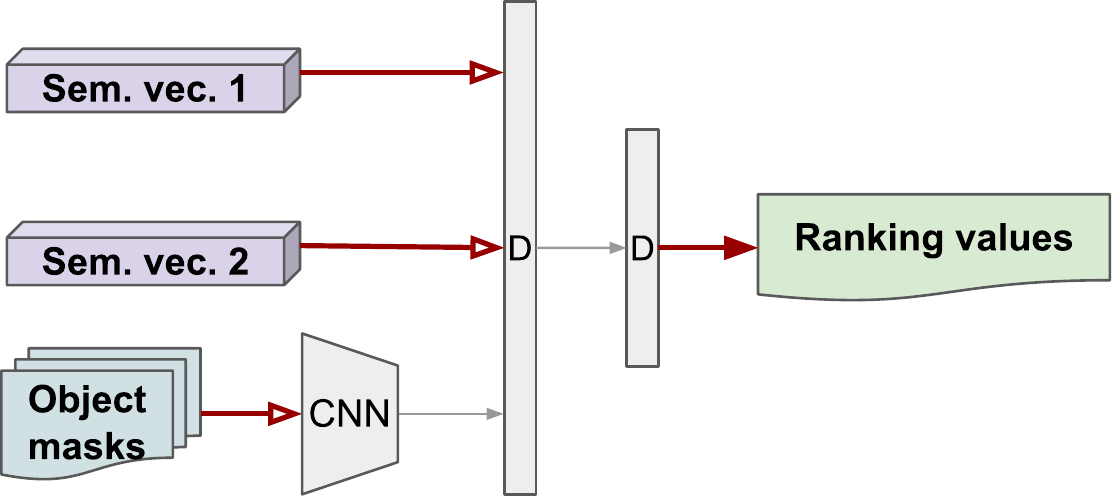}

	\caption{\textbf{Left: Simple Semantic model}, only containing two fully connected layers (marked with D).
	\textbf{Right: Simple Semantic-Positional model}, also containing a CNN module for
	processing the object position masks.}
	\label{fig:simplemodels}
\end{figure*}

We decided to test OG-SGG with a different scene graph generation network, in order to further
exemplify its model agnosticity.
Specifically we prepared and tested two simple models (see Fig.~\ref{fig:simplemodels})
while reusing the existing training pipeline:

\begin{itemize}
	\item \textbf{Simple Semantic model}: The two semantic vectors are concatenated and fed to a
	hidden fully connected layer of size $800$, followed by ReLU activation, batch normalization
	and a final fully connected layer of size $N$ (number of relationship predicates to be detected).
	The intention behind this model is focusing
	purely on language priors between object classes and relation classes; while ignoring all
	other inputs (i.e. object locations, image features, etc.).
	\item \textbf{Simple Semantic-Positional model}: In addition to the above, an object mask processor component
	based on several CNN layers generates a new vector of the same size as the semantic vectors,
	which is concatenated alongside the two semantic vectors (forming the input of the hidden layer).
	The hidden layer is also expanded to $1200$ neurons. The intention behind this model is
	augmenting the Simple Semantic model with object location information.
\end{itemize}

Table~\ref{table:semonlymodelteresa} shows the Simple Semantic model evaluated on TERESA,
Table~\ref{table:semonlymodelai2thor} shows the Simple Semantic model evaluated on AI2THOR, and
Table~\ref{table:semposmodelai2thor} shows the Simple Semantic-Positional model evaluated on AI2THOR.
The first thing to note is the performance uplift delivered by the post-processing
logic, further proving the usefulness of enforcing ontological axioms on generated scene graphs.
In the TERESA dataset, the combined efforts of the filtering and the post-processing are able to
easily outperform the baseline. Applying filtering without post-processing also generally produces
improved results, however there are regressions in some \textbf{R@K} metrics with low \textbf{K}
and \textbf{k} -- we believe this to be a side effect of increasing generalization capability by
reducing overfitting. In the case of AI2THOR, both simple models have trouble learning useful
information through the training process. As expressed in Section~\ref{sec:ai2thor}, this may be due to the harsher
filtering caused by the AI2THOR ontology. Nevertheless, the Simple Semantic model is able to dominate in
\textbf{mR@K}, indicating greater generalization capability. Interestingly, even though the Simple Semantic-Positional model
performs better in its baseline without OG-SGG than the Simple Semantic model, this initial advantage is unable
to materialize into improved across-the-board performance with full OG-SGG applied,
only managing to improve when post-processing without a filtered dataset is used.

Another comparison was made between the OG-SGG enriched Simple Semantic model and the baseline performance
of a non-trivial model (VRD-RANS). Despite having an order of magnitude fewer parameters
(VRD-RANS: 25M parameters, Simple Semantic model: 1.3M parameters), the obtained performance clearly
exceeds that of baseline VRD-RANS without OG-SGG. The same can be said for the Simple Semantic-Positional model,
however the gap in parameter count (25M vs 9.2M) is smaller.

\begin{table*}[tb]
	\centering
	\insetfont
	\caption{TERESA evaluation results on \textbf{Simple Semantic} model. The model was trained on
	4 different dataset splits (2 for each source training dataset), and evaluated
	with and without post-processing (\textit{``Post''} column). The baseline results for
	VRD-RANS trained on both source datasets are also shown in the table.}
	\begin{tabular}{ |l|c|c c c|c c c||c c c|c c c| }
		\cline{3-14}
		\multicolumn{2}{c|}{} & \multicolumn{12}{c|}{Metrics for Predicate Detection (PredDet)} \\
		\hline
		\multicolumn{1}{|c|}{\multirow{2}{*}{Dataset}} & \multirow{2}{*}{Post}
			& \multicolumn{3}{c|}{\textbf{R@K} (\textbf{k} = 1)} & \multicolumn{3}{c||}{\textbf{R@K} (\textbf{k} = 8)} & \multicolumn{3}{c|}{\textbf{mR@K} (\textbf{k} = 1)} & \multicolumn{3}{c|}{\textbf{mR@K} (\textbf{k} = 8)} \\
			& & 20 & 50 & 100 & 20 & 50 & 100 & 20 & 50 & 100 & 20 & 50 & 100 \\
		\hline
		\multirow{2}{*}{VG-SGG unmodified dataset}
			& \N & 22.9 & 31.3 & 40.1 & 20.0 & 26.9 & 39.8 & 19.4 & 29.5 & 39.1 & 25.6 & 36.1 & 47.6 \\
			& \Y & 26.0 & 33.6 & 44.0 & 26.8 & 36.2 & 52.0 & 34.5 & 43.3 & 54.3 & 30.3 & 44.3 & 61.2 \\
		\hline
		\multirow{2}{*}{VG-SGG with filtering/augmentation}
			& \N & 21.5 & 25.0 & 35.8 & 19.4 & 31.2 & 42.8 & 31.3 & 40.2 & 45.6 & 31.9 & 42.5 & 47.9 \\
			& \Y & \textbf{36.2} & \textbf{46.0} & \textbf{54.2} & \textbf{36.1} & \textbf{47.7} & \textbf{58.2} & \textbf{37.6} & \textbf{52.3} & \textbf{65.0} & \textbf{37.4} & \textbf{54.1} & \textbf{68.4} \\
		\hline
		\hline
		\multirow{2}{*}{VG-indoor unmodified dataset}
			& \N & 26.8 & 33.1 & 40.3 & 24.6 & 33.2 & 45.9 & 12.7 & 19.5 & 25.3 &  9.9 & 29.6 & 42.7 \\
			& \Y & 28.0 & 33.8 & 41.6 & 28.5 & 39.7 & 51.6 & 18.7 & 29.1 & 39.3 & 23.3 & 37.9 & 50.8 \\
		\hline
		\multirow{2}{*}{VG-indoor with filtering/augmentation}
			& \N & \textbf{34.8} & 36.3 & 42.1 & 29.4 & 39.7 & 46.6 & 27.9 & 36.4 & 52.3 & 24.8 & 37.9 & 52.2 \\
			& \Y & 34.5 & \textbf{37.9} & \textbf{45.3} & \textbf{36.7} & \textbf{42.4} & \textbf{55.7} & \textbf{28.0} & \textbf{37.4} & \textbf{53.4} & \textbf{29.1} & \textbf{39.8} & \textbf{62.0} \\
		\hline
		\multicolumn{14}{c}{} \\
		\hline
		\rowcolor{TableGray}
		\textit{VG-SGG unmodified (VRD-RANS)}
			& \N & 27.0 & 34.7 & 41.9 & 23.8 & 34.8 & 51.1 & 19.1 & 30.6 & 36.4 & 29.3 & 42.7 & 57.0 \\
		\hline
		\rowcolor{TableGray}
		\textit{VG-indoor unmodified (VRD-RANS)}
			& \N & 26.1 & 33.7 & 40.2 & 26.0 & 41.4 & 56.1 & 10.5 & 20.6 & 25.2 & 19.8 & 35.5 & 53.6 \\
		\hline
	\end{tabular}
	\label{table:semonlymodelteresa}
\end{table*}

\begin{table*}[tb]
	\centering
	\insetfont
	\caption{AI2THOR evaluation results on \textbf{Simple Semantic} model. The model was trained on
	4 different dataset splits (2 for each source training dataset), and evaluated
	with and without post-processing (\textit{``Post''} column). The baseline results for
	VRD-RANS trained on both source datasets are also shown in the table.}
	\begin{tabular}{ |l|c|c c c|c c c||c c c|c c c| }
		\cline{3-14}
		\multicolumn{2}{c|}{} & \multicolumn{12}{c|}{Metrics for Predicate Detection (PredDet)} \\
		\hline
		\multicolumn{1}{|c|}{\multirow{2}{*}{Dataset}} & \multirow{2}{*}{Post}
			& \multicolumn{3}{c|}{\textbf{R@K} (\textbf{k} = 1)} & \multicolumn{3}{c||}{\textbf{R@K} (\textbf{k} = 8)} & \multicolumn{3}{c|}{\textbf{mR@K} (\textbf{k} = 1)} & \multicolumn{3}{c|}{\textbf{mR@K} (\textbf{k} = 8)} \\
			& & 20 & 50 & 100 & 20 & 50 & 100 & 20 & 50 & 100 & 20 & 50 & 100 \\
		\hline
		\multirow{2}{*}{VG-SGG unmodified dataset}
			& \N & 14.1 & 18.5 & 24.9 & 15.9 & 20.4 & 33.3 &  4.5 &  7.5 & 10.5 &  6.7 & 11.1 & 16.9 \\
			& \Y & \textbf{17.3} & \textbf{25.6} & \textbf{36.3} & \textbf{18.0} & \textbf{26.1} & \textbf{42.9} &  5.2 &  9.3 & 14.0 &  \textbf{7.7} & 13.0 & 21.7 \\
		\hline
		\multirow{2}{*}{VG-SGG with filtering/augmentation}
			& \N &  4.8 & 10.3 & 14.7 &  6.8 & 14.9 & 24.9 &  5.8 &  9.6 & 13.6 &  6.0 & 12.0 & 20.3 \\
			& \Y &  5.0 & 11.4 & 18.0 &  8.3 & 17.7 & 32.4 &  \textbf{5.9} &  \textbf{9.9} & \textbf{15.5} &  7.0 & \textbf{13.2} & \textbf{26.1} \\
		\hline
		\hline
		\multirow{2}{*}{VG-indoor unmodified dataset}
			& \N & 17.0 & 19.6 & 25.5 & 19.0 & 24.0 & 34.8 &  7.1 & 11.9 & 17.0 &  8.5 & 17.6 & 26.1 \\
			& \Y & \textbf{19.4} & \textbf{25.0} & \textbf{32.9} & \textbf{21.2} & \textbf{28.7} & 42.4 &  7.8 & 14.9 & 20.7 &  9.3 & \textbf{20.1} & 29.7 \\
		\hline
		\multirow{2}{*}{VG-indoor with filtering/augmentation}
			& \N & 16.6 & 21.7 & 27.8 & 18.0 & 23.8 & 36.2 &  9.5 & 15.0 & 19.7 & 10.2 & 18.6 & 28.7 \\
			& \Y & 17.3 & 24.6 & 32.7 & 19.0 & 28.1 & \textbf{43.3} &  \textbf{9.8} & \textbf{15.9} & \textbf{21.3} & \textbf{10.4} & 20.0 & \textbf{32.1} \\
		\hline
		\multicolumn{14}{c}{} \\
		\hline
		\rowcolor{TableGray}
		\textit{VG-SGG unmodified (VRD-RANS)}
			& \N & 19.1 & 24.4 & 32.1 & 21.9 & 27.2 & 40.5 &  7.0 & 11.8 & 16.2 &  9.4 & 15.2 & 24.6 \\
		\hline
		\rowcolor{TableGray}
		\textit{VG-indoor unmodified (VRD-RANS)}
			& \N & 18.7 & 21.4 & 26.0 & 22.1 & 26.5 & 37.4 &  9.3 & 17.7 & 23.0 & 12.2 & 22.3 & 31.5 \\
		\hline
	\end{tabular}
	\label{table:semonlymodelai2thor}
\end{table*}

\begin{table*}[tb]
	\centering
	\insetfont
	\caption{AI2THOR evaluation results on \textbf{Simple Semantic-Positional} model. The model was trained on
	4 different dataset splits (2 for each source training dataset), and evaluated
	with and without post-processing (\textit{``Post''} column). The baseline results for
	VRD-RANS trained on both source datasets are also shown in the table.}
	\begin{tabular}{ |l|c|c c c|c c c||c c c|c c c| }
		\cline{3-14}
		\multicolumn{2}{c|}{} & \multicolumn{12}{c|}{Metrics for Predicate Detection (PredDet)} \\
		\hline
		\multicolumn{1}{|c|}{\multirow{2}{*}{Dataset}} & \multirow{2}{*}{Post}
			& \multicolumn{3}{c|}{\textbf{R@K} (\textbf{k} = 1)} & \multicolumn{3}{c||}{\textbf{R@K} (\textbf{k} = 8)} & \multicolumn{3}{c|}{\textbf{mR@K} (\textbf{k} = 1)} & \multicolumn{3}{c|}{\textbf{mR@K} (\textbf{k} = 8)} \\
			& & 20 & 50 & 100 & 20 & 50 & 100 & 20 & 50 & 100 & 20 & 50 & 100 \\
		\hline
		\multirow{2}{*}{VG-SGG unmodified dataset}
			& \N & 12.3 & 18.3 & 26.1 & 16.7 & 25.5 & 40.8 & 12.9 & 17.1 & 20.7 & 15.6 & 23.6 & 32.1 \\
			& \Y & \textbf{18.5} & \textbf{27.7} & \textbf{38.6} & \textbf{24.8} & \textbf{34.0} & \textbf{51.0} & \textbf{15.4} & 19.9 & 24.3 & \textbf{18.9} & 27.5 & 37.8 \\
		\hline
		\multirow{2}{*}{VG-SGG with filtering/augmentation}
			& \N &  9.3 & 17.7 & 24.7 &  8.4 & 17.9 & 33.4 &  9.9 & 18.3 & 23.6 &  9.3 & 22.6 & 40.3 \\
			& \Y & 10.7 & 20.6 & 29.1 & 11.1 & 25.0 & 43.8 & 11.6 & \textbf{20.4} & \textbf{26.0} & 11.6 & \textbf{27.8} & \textbf{48.2} \\
		\hline
		\hline
		\multirow{2}{*}{VG-indoor unmodified dataset}
			& \N & 21.5 & 24.0 & 27.1 & 23.5 & 28.6 & 38.5 & 10.6 & 18.4 & 21.7 & 11.7 & 22.9 & 31.6 \\
			& \Y & \textbf{24.8} & \textbf{29.6} & \textbf{36.7} & \textbf{26.5} & \textbf{33.2} & \textbf{45.4} & \textbf{12.6} & 20.7 & 25.4 & 13.0 & 25.5 & 36.6 \\
		\hline
		\multirow{2}{*}{VG-indoor with filtering/augmentation}
			& \N & 19.5 & 22.9 & 27.4 & 21.4 & 26.6 & 37.2 & 12.2 & 19.9 & 24.6 & 13.3 & 23.1 & 34.6 \\
			& \Y & 20.6 & 26.0 & 31.9 & 22.8 & 30.7 & 44.0 & \textbf{12.6} & \textbf{21.4} & \textbf{27.0} & \textbf{13.9} & \textbf{26.2} & \textbf{39.9} \\
		\hline
		\multicolumn{14}{c}{} \\
		\hline
		\rowcolor{TableGray}
		\textit{VG-SGG unmodified (VRD-RANS)}
			& \N & 19.1 & 24.4 & 32.1 & 21.9 & 27.2 & 40.5 &  7.0 & 11.8 & 16.2 &  9.4 & 15.2 & 24.6 \\
		\hline
		\rowcolor{TableGray}
		\textit{VG-indoor unmodified (VRD-RANS)}
			& \N & 18.7 & 21.4 & 26.0 & 22.1 & 26.5 & 37.4 &  9.3 & 17.7 & 23.0 & 12.2 & 22.3 & 31.5 \\
		\hline
	\end{tabular}
	\label{table:semposmodelai2thor}
\end{table*}

\section{Conclusion and future work}

In this work we joined the world of ontologies together with the world of
scene graph generation, and showed how the strategies we proposed
(using only filtering and processing based on the most common OWL axioms
affecting predicates)
can achieve quantitative and qualitative improvements in domain specific
environments. Whereas existing scene graph generation networks (such as
VRD-RANS) generate all possible pairs, the OG-SGG methodology is able to leverage the ontology
to reduce the set of possibilities and thus improve the quality of the generated
scene graphs. We can observe improvements across the board in \textbf{R@K}, and
interestingly enough, training the model with a smaller version of the dataset
resulted in improved \textbf{mR@K}, especially when compared to the baseline
(that is, the model trained on the original version of the dataset without OG-SGG).
Evaluating the performance without graph constraint priors (i.e. by setting
the graph constraint hyperparameter \textbf{k} to its highest allowed value)
also produced better results. We also show how OG-SGG improves the
results for different application scenarios (TERESA and AI2THOR datasets),
and also for different scene graph generation models.

Another important observation is that only a small amount of effort had to be
spent in engineering an ontology for the experiment in order to obtain these results.
Specifically, the only two things that need to be done for OG-SGG to work are
designing an ontology for the desired problem, and mapping the predicates of the
original scene graph dataset to the ones in the ontology.
It can be explained that OG-SGG leverages the effect that biased datasets
have on neural networks, precisely by creating a new version of the dataset
that is \textit{biased} in favor of existing prior knowledge. On the other
side of the equation, OG-SGG also removes outputs that can be safely discarded
using the aforementioned prior knowledge.
This contrasts with the traditional methods used for transfer learning in neural
networks, which are primarily based on hyperparameter tweaking, freezing and
unfreezing the weights of individual layers, and other ``black box''
architectural changes.
All of these methods, as with other non-XAI techniques, cannot be driven by
human intuition or by pre-existing knowledge; and as such take a considerably
higher amount of effort to refine, mostly through pure trial and error.

Nevertheless, there is margin for further refinements and
filtering. It still takes a high \textbf{K} cutoff to capture a sizable
majority of the triplets present in ground truth annotations, indicating
a need for better filtering.
The quality of the filtering also depends on how detailed the ontology
is -- naturally, the more axioms and predicates that exist the more precise the
predictions will be. In addition, a major flaw with existing scene graph
generation networks can be observed, which is the difficulty of defining
a score threshold for dropping unlikely relation triplets.
Currently, a basic Top-K strategy is still used, which tends to leave out
perfectly valid predictions in crowded scenes.
For this reason,
a possible future direction would be to design a new post processing stage
based on a neural network that draws the line for us, and possibly even go
further by filtering with (higher order) axioms from the ontology.
Another area of interest for possible future research is integrating
existing ontological knowledge directly into the main scene graph generation
network, perhaps in the form of a new term in the loss function
\cite{diaz2022explainable}, or through incorporating neurosymbolic
propositional and first order logic directly as part of the training
process \cite{badreddine2022logic}.
Simultaneous Localization and Mapping (SLAM) systems could be yet another area of
interest for future work related to ontology-guided machine learning.
Specifically,
Semantic SLAM systems capable of segmenting rooms and labelling/tracking all objects
within could be proposed, and this potentially involves the detection of relationships
between objects in a similar way to the scene graph generation problem.

On an ending note, we propose further research on downstream usages of
OG-SGG such as knowledge-graph driven image captioning or robotic visual
question answering, further leveraging structured approaches to incorporating prior relevant knowledge.

\subsection*{Acknowledgment}

This work is partially supported by Programa Operativo FEDER Andalucia
2014-2020, Consejeria de Economía y Conocimiento (TELEPORTA, UPO-1264631 and DeepBot, PY20\_00817)
and the project PLEC2021-007868, funded by MCIN/AEI/10.13039/501100011033
and the European Union NextGenerationEU/PRTR.
N. Díaz-Rodríguez is supported by the Spanish Government Juan de la Cierva
Incorporación contract (IJC2019-039152-I) and Google Research Scholar Programme.

\bibliographystyle{plain}
\footnotesize{ \bibliography{OGSGG}{} }

\end{document}